\documentclass[10pt,onecolumn,times,review]{elsarticle}
\usepackage{graphicx}
\usepackage{epstopdf}
\usepackage{natbib}
\usepackage{amsmath,amssymb,amsfonts}
\usepackage{algorithmic}
\usepackage{mathptmx}
\usepackage{amsfonts}
\usepackage{algorithm}
\usepackage{multirow}
\usepackage{dcolumn}
\usepackage{color, colortbl}

\usepackage{mathtools}

\let\vec\mathbf

\definecolor{Gray}{gray}{0.9}

\journal{arXiv}

\begin{document}

\begin{frontmatter}


\title{Evolving Multi-Label Fuzzy Classifier\footnote{This paper was accepted and published in {\em Information Sciences}: https://doi.org/10.1016/j.ins.2022.03.045}}        

\author[KBMS]{Edwin Lughofer}
\ead{edwin.lughofer@jku.at (Corresponding Author)}

\address[KBMS]{Institute for Mathematical Methods in Medicine and Data Based Modeling, Johannes Kepler University Linz, Austria}






\begin{abstract}
Multi-label classification has attracted much attention in the machine learning community to address the problem of assigning single samples to more than one (not necessarily non-overlapping) class at the same time. We propose an evolving multi-label fuzzy classifier (EFC-ML) which is able to self-adapt and self-evolve its structure with new incoming multi-label samples in an incremental, single-pass manner. It is based on a multi-output Takagi-Sugeno type architecture, where for each class a separate consequent hyper-plane is defined, which yields flexibility for partially approximating the respective classes in a binary $[0,1]$-regression context. The learning procedure embeds a locally weighted incremental correlation-based algorithm combined with (conventional) recursive fuzzily weighted least squares and Lasso-based regularization. Locality is important to avoid the out-masking effect of single class labels in one or more rules; the correlation-based part ensures that the interrelations between class labels, a specific well-known property in multi-label classification for improved performance, are preserved properly; the Lasso-based regularization reduces the curse of dimensionality effects in the case of a higher number of inputs. Antecedent learning is achieved by product-space clustering and conducted for all class labels together, which yields a single rule base (opposed to related techniques such as one-versus-rest or classifier chaining, achieving multiple different rule bases, one per class), allowing a compact knowledge view and thus enabling better interpretable insights.
Furthermore, our approach comes with an online active learning (AL) strategy for updating the classifier on just a (smaller) number of selected samples, which in turn makes the approach applicable for scarcely labelled streams in applications, where the annotation effort is typically expensive. It is based on three essential concepts: novelty content in the antecedent space, uncertainty due to ambiguity in the consequent (output) space and parameter instability reduction, and these in combination with an upper-allowed selection budget (which could be predefined by a user).
Our approach was evaluated on several data sets from the MULAN repository and showed significantly improved classification accuracy and average precision trend lines compared to (evolving) one-versus-rest or classifier chaining concepts. significant result was that, due to the online AL method, a 90\% reduction in the number of samples used for classifier updates had little effect on the accumulated accuracy trend lines compared to a full update in most data set cases.
\end{abstract}
\begin{keyword}
data stream classification, evolving fuzzy classifiers, multi-labeling, incremental correlation-based learning, on-line active learning, sample selection
\end{keyword}

\end{frontmatter}

\section{Introduction}
\label{introduction}
Due to the increasing complexity, dynamics and non-stationarity of current industrial installations \cite{LughoferMouchaweh19}, data-driven machine learning models have to meet the flexibility of online processing demands in order to maintain their predictive accuracy or even improve it based on newly recorded (online) samples, e.g., representing new operation modes, system states or changing environmental behavior \cite{Angelov12}.
In terms of classification problems often used for (complex) decision-making processes, such as anomaly and fault detection \cite{BehniafarNowrooziShahriari18} or predictive and preventive maintenance operations \cite{LughoferMouchaweh19}, incremental learning techniques, which are able to update classifiers on demand and on the fly, have been proposed in the literature \cite{HisadaOzawaZhangKasabov10}.

Evolving classifiers and in particular evolving (neuro-)fuzzy classifiers (EFCs) are a prominent field of research on incremental learning algorithms, as they i.) are able to self-evolve their structure from newly incoming (stream) samples, allowing them to incrementally expand their knowledge to new systems states \cite{LughoferChapter14} \cite{SkrjancIglesiasLughoferGomide19}, newly arising classes, novelty content in stream samples \cite{AngelovFilevKasabov10}, drifts \cite{PratamaAnavattiLu15} etc.; ii.) are able to address uncertainty in the data (e.g., due to measurement noise) in a natural way through the concepts of granules \cite{GarciaLeiteSkrjanc20} \cite{LeiteAndonovskiSkrjancGomide19}, typically realized through fuzzy sets and rules \cite{PedryczSkowronKreinovich08}; and iii.) allow some sort of interpretability and thus insights into the process due to the extraction of linguistically readable IF-THEN rules, which also offers good explainability of model classification outputs  \cite{LughoferRichterNeisslHeidlEitzingerRadauer17}.
The first aspect is necessary to properly deal with the increasing dynamics and speed of today's data streams.
The latter two aspects are not so easily and directly addressable with other types of ML models such as extreme learning machines, (deep) neural networks, support vector machines etc., and specific sophisticated techniques have to be integrated to make such ML models interpretable and their outputs explainable \cite{SamekMontavon19}.


\subsection{Related State-of-the-Art and Motivation}
\label{motivation_SoA}
Several evolving fuzzy classifier approaches have been proposed in the literature in the past, with the first approach in \cite{AngelovLughoferZhou09} discussing different possible fuzzy classifier architectures (based on a single model, multi model, one-versus-rest techniques and beyond) to address (online) multi-class classification problems in an incremental and single-pass processing manner.
One key issue was to resolve the masking effect in the case of multi-class problems by a specific fuzzy regression scheme based on indicator vectors, leading to multiple classification models, i.e. one classifier per class.
The multi-model aspect of rule-based classifiers was further expanded in \cite{AngelovZhou08} and in several forms in \cite{PratamaAnavattiLughoferjour14} to address rule-based multi-model concepts rather than sole global model-based ones. The approach in \cite{LughoferBuchtala13} employed the all-pairs/pairwise classification concept to evolve (smaller) binary fuzzy classifiers for each class pair and to produce an overall classification output based on a preference relation matrix among the class pairs. It improved one-versus-rest and single EFCs in terms of predictive accuracy, especially for larger-scale multi-class classification problems embedding a higher number of classes and typically a significant class imbalance.

An extended evolving learning procedure for classification rules was proposed in \cite{KanginAngelovIglesias15}, which is based on the AnYa model architecture \cite{AngelovYager12} and extracts clouds as granules instead of fuzzy sets and rules with pre-defined distributions and shapes. It is also embedded in the ALMMo approach for learning multi-model event systems\cite{AngelovGuPrincipe18}.
A further remarkable approach was proposed in \cite{WangJiJin13}, where the linear weights of a classifier are updated based on a constrained optimization problem with the use of fuzzy membership degrees obtained through center-distance and margin-distance based assignments.

Extensions to integrate type-2 fuzzy sets to integrate the spirit of type-2 fuzzy classifiers \cite{JohnHagrasCastillo19} and to model a second level of uncertainty in evolving fuzzy classifiers were proposed in \cite{PratamaLuZhang16}, \cite{TungQuekGuan13} and \cite{SubramanianSureshSundararajan13}, the latter acting on three meta-cognitive aspects such as how-to-learn, when-to-learn and what-to-learn to maximize efficiency of the updates, the former exploiting the concept of interval-valued type-2 fuzzy sets in combination with Chebyshev polynomials in the rule consequents in order to locally model possible non-linearity. In
\cite{PratamaLuLughoferZhangAnavatti16}, an approach was demonstrated to integrate scaffolding aspects to react to drifts effectively (with short delays), which was extended in \cite{ZainLughoferPratama17} to mine web news articles effectively and put them into respective categories.


More recently, evolving neuro-fuzzy classifiers in classical \cite{SouzaLughoferNC20} and in deep rule-based form \cite{AngelovGu18} have attracted the attention of the research community. The former type of classifiers uses a classical neuro-fuzzy systems architecture (embedding a fuzzy set layer, a rule layer and a consequent layer), which expands to a multi-output model in the case of multi-class classification problems, where each output represents one class independently. Applying a softmax operator (as commonly used in neural classifiers) on the certainty vector delivers the final output class. Learning is typically conducted by a one-versus-rest classification technique through a loss function on each class (versus the rest) separately (see \cite{SouzaLughoferNC20}). Deep rule-based classifiers, recently proposed in \cite{AngelovGu18}, contain massively parallel ensemble of IF-THEN rules in the third layer as an essential core component. It is significant that the rule appearance in this architecture realizes OR-connections to describe a single class through the possibilities of prototypes (learned from the data based on empirical data analysis without assuming any specific data distribution \cite{AngelovGuPrincipe17}), which makes them nicely interpretable while allowing high predictive precision due to the deep (multiple rule) layer aspect.

The common denominator is that all of these EFC techniques were developed for the purpose of either binary or multi-class classification problems. No EFC approach for {\em multi-label} classification has been proposed so far. As will be explained in more detail in Section \ref{classifier_architecture}, multi-label classification differs from {\em multi-class} classification, as it allows a vector of possible class labels to which a sample belongs (e.g., there can be overlapping or nested groups to which a sample may belong, such as a bull terrier belongs to the classes 'dog' and 'animal', but not to the classes 'cat' or 'ball'). In general, in multi-label problems, there is no constraint as to how many of the (possible) classes to which a sample can be assigned \cite{HerreraCharte18} (also see Section \ref{classifier_architecture}). In this sense, multi-label classification has become more and more an important factor in the increasing complexity of todays’ classification systems \cite{ZhangZhou14}, typically inducing higher class ambiguities or widening the possible groups of classes to which samples may belong. This is even more the case within an online stream classification context \cite{HerreraCharte18}, for which an evolving fuzzy approach is desirable to meet the challenges discussed in the first two paragraphs above.


\subsection{Our Approach}
\label{our_approach}
We propose a new evolving fuzzy classifier designed for (online) multi-label classification problems, which does not simply perform learning on the various labels independently, but also takes into account possible interrelations among them, which is a specific property in multi-label classification \cite{HerreraCharte18}.
In summary, our approach is driven by the following aspects:
\begin{itemize}
\item The fuzzy classifier structure is based on the classical Takagi-Sugeno model architecture, but using multiple consequent functions per rule, one for each class label; each consequent function thus represents a regressor tendency for each label. In this sense, a multiple output TS fuzzy classifier is achieved whose continuous outputs are converted to a crisp classification response for each label (1 when the sample belongs to the respective class, 0 otherwise) through an indicator function.
\item A joint antecedent space among all class labels is incrementally learned from the data through an evolving clustering algorithm in the product-space, embedding rule evolution and merging concepts; this yields one compact rule base for the whole classification problem (instead of generating different rule bases for different class labels --- as is done in one-versus-rest or classifier chaining techniques), which leads to more transparent fuzzy classifiers.
\item The learning of consequent parameters is performed on a local level per rule separately to inherit its favorable properties regarding faster computation times, more robust updates etc. compared to global learning (as discussed in detail in \cite{AngelovZhou08}). This is achieved by a combination of recursive fuzzily weighted least squares \cite{LughoferChapter14} \cite{SkrjancIglesiasLughoferGomide19} and an incremental local correlation-based learning scheme, which takes into account possible locally weighted correlations among the labels and tries to preserve them in the regression coefficients --- this is motivated from past studies \cite{JiTangYuYe08} \cite{LiWangPavluAslam16}, where correlation-based learning has shown improved performance for multi-label classification in general (batch case). Furthermore, an additional regularization term on the parameters is integrated based on $L_1$ norm to shrink as many parameters as possible towards 0, which in turn may reduce the possible curse of dimensionality effects.
\item An online active learning scheme in connection with the evolving multi-label classifier architecture is proposed to meet the label scarcity issue due to unrealistic, high annotation efforts to provide the full class label information on all new incoming samples. It is based on the use of three sample selection criteria, which check the degree of novelty content, the degree of uncertainty in model outputs as well as the degree of parameter stability increase of a new sample for the current fuzzy classifier. Furthermore, it embeds a budget-based learning technique to stay inside an allowed maximal percentage of selected samples.
\end{itemize}
We evaluated our approach on several multi-label classification data sets available in the well-known MULAN repository\footnote{http://mulan.sourceforge.net/datasets-mlc.html} (see Section \ref{experimental_setup}), with a varying number of class labels, attributes and number of samples, and compared it with related multi-label concepts such as classifier chaining \cite{ReadPfahringerHolmes11} and one-versus-rest classification \cite{AngelovZhou08}, both operating also in an incremental and evolving manner based on fuzzy classifier architecture. This provided us a fair comparison as to how the multi-output architecture and especially the incremental (locally) weighted and regularized correlation-based learning could improve the classification performance (see Section \ref{results}). Furthermore, we applied our online active learning technique to check to what degree a reduction in the requested sample annotations impacts classification accuracy. A significant result was that a reduction down to 10\% had a minimal effect on the accumulated (ahead prediction) accuracy trend lines in most data set cases. In the subsequent section, we describe the problem and the classifier architecture, which is followed by a comprehensive section about the methodological aspects of our approach.


\section{Problem Statement and Classifier Architecture}
\label{classifier_architecture}
In this section, we define the problematic issue of multi-label classification and suggest a reliable fuzzy classification model to deal with it.
First of all, {\em multi-label} classification should not be confused with {\em multi-class} classification. The latter relies on the existence of several classes in a classification problem and associated data set, but for each sample, a unique class can be assigned. For instance, the well-known iris flower data set is a typical multi-class classification problem, as no samples from flowers crossings are included (no mixed cultures).
Multi-label classification problems, however, induce that samples may belong to more than one class at the same time (but not necessarily).
For instance, in a classification problem where the classes are dog, animal and ball, a bull terrier would belong to the classes 'dog' AND 'animal', but not to the class 'ball' (two-class assignment), whereas a monkey would belong to 'animal', but not to 'dog' and 'ball' (single class assignment).

Thus, formally speaking, a multi-label classification data set is defined by $(X,Y)$, where $X\in \mathbb{R}^{N\times p}$ denotes a (classical) input feature matrix with $p$ features and $N$ samples and $Y$ denotes a label structure which can be a set of classes for each sample (row) $\vec{x}\in X$, indicating to which classes the corresponding samples belong. As the number of classes can be different for different samples in $X$, this leads to a different output dimensionality, which is typically not easy to handle in data-driven algorithms. We can simply remove this by defining $Y\in \mathbb{R}^{N\times K}$ as an indicator matrix, with $K$ number of classes, where each of the $K$ entries is either 0 or 1, depending on whether the corresponding sample (row) belongs to the respective class or not.
For instance, in the dog-animal-ball example above, the sample of a bull terrier would receive a multi-label vector of $\vec{y} = \{1,1,0\}$, whereas the sample of a monkey would receive a multi-label vector of $\vec{y} = \{0,1,0\}$.

As each column vector in $Y$ denotes an indicator vector where the positions of the $1$'s indicate which samples belong to the corresponding class, it is intuitive to extend the consequent space of a conventional fuzzy system to a multiple output space. In other words, each rule in a fuzzy classifier has a set of consequences, each one belonging to one class. Whenever a particular rule has a consequent for the $k$th class, which approaches 1, it thus indicates that this rule represents the $k$th class (but it may also represent other classes synchronously due to its multiple consequents $\rightarrow$ multi-label rule).
Formally, a fuzzy rule $R_i$ with $K$ multiple consequences (for $K$ classes) is defined as:
\begin{align}
\label{fuzzyrule_multilabel_classifier}
R_i: \hspace{0.1cm} & \text{IF} \hspace{0.1cm} x_1 \hspace{0.1cm} \text{IS}
\hspace{0.1cm} \mu_{i1} \hspace{0.1cm} \text{AND}
... \text{AND} \hspace{0.1cm} x_p \hspace{0.1cm} \text{IS}
\hspace{0.1cm} \mu_{ip} \hspace{0.1cm} \nonumber \\
& \text{THEN} \hspace{0.1cm} \vec{l}_{i}(\vec{x}) = \left[ \begin{array}{c}
l_{i,1}(\vec{x})   \\
l_{i,2}(\vec{x})   \\
\vdots      \\
l_{i,K}(\vec{x})  \\
\end{array}
\right] = \left[ \begin{array}{c}
w_{i0,1} + w_{i1,1}x_1 + ... + w_{ip,1}x_p   \\
w_{i0,2} + w_{i1,2}x_1 + ... + w_{ip,2}x_p   \\
\vdots      \\
w_{i0,K} + w_{i1,K}x_1 + ... + w_{ip,K}x_p  \\
\end{array}
\right]
\end{align}
with $p$ the input dimensionality of the classification problem and $\mu_{ij}$ the $j$th fuzzy set in the $j$ antecedent part of the $i$th rule.


Here, we use hyper-planes as consequents (instead of singletons indicating direct class membership) to allow membership to a class in dependency of the input features for a particular rule. This typically achieves more accurate classification performance, as thoroughly studied before, e.g. in \cite{AngelovLughoferZhou09}. 
Please note that the antecedent part for each rule is the same for all classes, i.e. one joint antecedent space over all classes (and multi-label combinations) is represented for the whole rule base. This makes knowledge interpretation through the rules compact and easy to understand. Furthermore, it differs from the one-versus-rest classification technique as shown in \cite{AngelovLughoferZhou09}, which uses the same indicator entries for the consequents, but learns a different model and thus a different antecedent space for different classes.

The inference to produce a continuous multi-label classification output for a new query sample $\vec{x}$ is calculated by:
\begin{equation}
\label{multilabel_classifiers_inference}
\vec{f}(\vec{x}) = \hat{\vec{y}} = \sum_{i=1}^{C} \Psi_i(\vec{x}) \cdot \vec{l}_{i}(\vec{x}) = \left[ \begin{array}{c}
\sum_{i=1}^{C} \Psi_i(\vec{x}) \cdot l_{i,1}(\vec{x})   \\
\sum_{i=1}^{C} \Psi_i(\vec{x}) \cdot l_{i,2}(\vec{x})   \\
\vdots      \\
\sum_{i=1}^{C} \Psi_i(\vec{x}) \cdot l_{i,K}(\vec{x})  \\
\end{array}
\right]
\hspace{0.3cm} \Psi_i(\vec{x}) =
\frac{\mu_i(\vec{x})}{\sum_{j=1}^{C} \mu_j(\vec{x})},
\end{equation}
where $\mu_i(\vec{x})$ denotes the rule activation (membership) degree of $\vec{x}$ to the $i$th rule, and is given by:
\begin{equation}
\label{multivariate_Gaussian_distribution}
\mu_i(\vec{x}) = \exp (-\frac{1}{2} (\vec{x}-\vec{c}_i)^T\Sigma_i^{-1}(\vec{x}-\vec{c}_i))
\end{equation}
with $\vec{c}_i$ the center and $\Sigma_i^{-1}$ the inverse covariance matrix of the $i$th rule (which defines its orientation and shape).
The weights, given by $\Psi_i(\vec{x})$, represent the normalized rule fulfillment degrees in sample $\vec{x}$, i.e. the membership degrees in each rule in relation to the sum of memberships in all rules. Therefore, the rules lying closer to a current sample than other rules receive a higher weight, thus their local linear trends described by $l_{i,.}$ are more active. This also yields a kind of local interpretation (per rule) of the impact of features for approximating the classes (through their regression coefficients $\vec{w}$ in the respective class consequents) and can be used for explainability concepts \cite{LughoferRichterNeisslHeidlEitzingerRadauer17}.

$\hat{\vec{y}}$ in \eqref{multilabel_classifiers_inference} contains one (continuous) prediction value for each class, which is then thresholded in the following way to obtain the crisp multi-class output:
\begin{equation}
\label{multilabel_output}
\hat{\vec{y}}_c = \mathbb{I}_{\hat{\vec{y}}_j \geq 0.5}, \hspace{0.1cm} j = 1,...,K
\end{equation}
where $\mathbb{I}$ denotes a classical indicator function, i.e. which delivers 1 when the condition in the subscripting index is fulfilled, and 0 otherwise.

Please note that the definition of rule activation levels through \eqref{multivariate_Gaussian_distribution} yields the generalized form with ellipsoidal shapes in an arbitrarily rotated position --- whenever $\Sigma_i^{-1}$ approaches a diagonal matrix, the classical, axis-parallel form is achieved and the rules are then given as defined in \eqref{fuzzyrule_multilabel_classifier}.
Even if $\Sigma_i^{-1}$ is a general inverse covariance matrix, linguistically readable rules in IF-THEN form with AND-connections as defined in \eqref{fuzzyrule_multilabel_classifier} can be achieved due to projection concepts as discussed in \cite{LughoferCernudaKindermannPratama14}.


\section{Methodology for Online Classifier Training and Evolution}
\label{methodology}
We present the core aspects of our incremental, evolving learning policy for multi-label fuzzy classifiers. The main meat is contained in the new recursive learning technique for the consequent parameters to exploit correlations between class labels, which is typically a peculiar characteristic of multi-label classification problems \cite{ZhangYeung13} and usually increases the accuracy of the classifiers, as shown in \cite{JiTangYuYe08} \cite{LiWangPavluAslam16}.
\subsection{Antecedent Learning}
\label{antecedent_learning}
As our multi-label fuzzy classifier architecture contains a joint antecedent space for all classes, this can be learnt in an online, incremental manner on data stream samples by any classical evolving rule extraction approach typically realized through evolving clustering processes (see \cite{SkrjancIglesiasLughoferGomide19} for a comprehensive survey). Whenever a new cluster is evolved subject to the fulfillment of a specific rule evolution criterion, automatically a new rule is born with multiple consequents (one per class), and its parameters are typically set to the nearest rule or to a vector containing just 0's to ensure fast convergence \cite{AngelovFilev04}.
Due to past experience, we applied our own evolving rule extraction technique based on generalized clusters, as proposed in \cite{LughoferCernudaKindermannPratama14}.

One essential aspect is the rule evolution criterion which is responsible for deciding when to evolve a new rule on demand. This criterion will also be used for checking the novelty content of new samples with respect to already existing clusters in the active learning component (see Section \ref{online_activelearning}). It is defined based on statistically funded tolerance regions of multivariate Gaussians as induced by \eqref{multivariate_Gaussian_distribution}:
\begin{align}
\label{rule_evolution_criterion}
& min_{i=1,...,C} \hspace{0.2cm} (ma_i = \sqrt{(\vec{x}-\vec{c}_i)^T \Sigma_i^{-1} (\vec{x}-\vec{c}_i)}) > r_{win}, \hspace{0.2cm} r_{win} = fac*p^{1/\sqrt{2}} \frac{1.0}{\left(1-1/(k_{win}+1)\right)^m}
\end{align}
with $win$ denoting the rule index with minimal $ma_i$; $\vec{x}$ denotes the current stream sample, $\vec{c}_i$ the center of the $i$th rule and $\Sigma_i^{-1}$ its inverse covariance matrix; $k_i$ denotes the number of samples supporting the $i$th rule, $p$ the input dimensionality and $m$ is a constant factor which is set to 4 per default; $fac$ is an a priori defined parameter, steering the tradeoff between stability (update of an old cluster) and plasticity (evolution of a new cluster). This is the only sensitive parameter in our method and can be optimized on an initial batch data set (e.g., on the first portion of a stream).
If this condition is fulfilled, a new rule is evolved by setting its center $\vec{c}_{C+1}$ to the current sample $\vec{x}$ and its inverse covariance matrix $\Sigma_i^{-1}$ to a diagonal matrix containing a vector $[1/(\vec{eps}*\vec{\sigma})]$, with $\vec{eps}$ a vector containing a small fraction of the ranges of the features, where $/$ and $*$ denote component-wise division and multiplication, respectively. If \eqref{rule_evolution_criterion} is not fulfilled, the inverse covariance matrix and the center of the nearest rule (with index $win$) are updated, see \cite{LughoferCernudaKindermannPratama14}.

A particular consideration for the multi-label case is whether to perform evolving clustering solely in the input space or in the product space.
We choose the latter to trigger new clusters for new classes (i.e. new $\{0,1\}$-combinations), thus the input on our incremental learning procedure is the sample with its complete indicator vector, thus $(\vec{x},\vec{y})$.

\subsection{Consequent Learning}
\label{consequent_learning}
\subsubsection{Optimization Function Formulation}
The objective function for establishing consequent learning of the linear parameters $W$, collected in a matrix with columns denoting the single labels and with rows denoting the hyper-planes of all of the rules together (see also \eqref{fuzzyrule_multilabel_classifier}), contains three terms: the regression loss term, the regularization term and the correlation-based learning term to take into account the possible correlations among the classes.
These three terms are localized, to define a weighted objective function per rule separately and to induce a separate estimation of the consequent parameters per rule. This is due to various intrinsic studies in the past (see, e.g. \cite{AngelovLughoferZhou09} \cite{LughoferPositionPaper13}), which made clear that local consequent learning induces smaller matrices to invert and thus is more robust and quicker to calculate solutions which are also more interpretable, as the hyper-planes locally snuggle along the final approximation surfaces \cite{LughoferPositionPaper13}.

\begin{figure}[t]
\centering
\fbox{\includegraphics[width=0.65\textwidth]{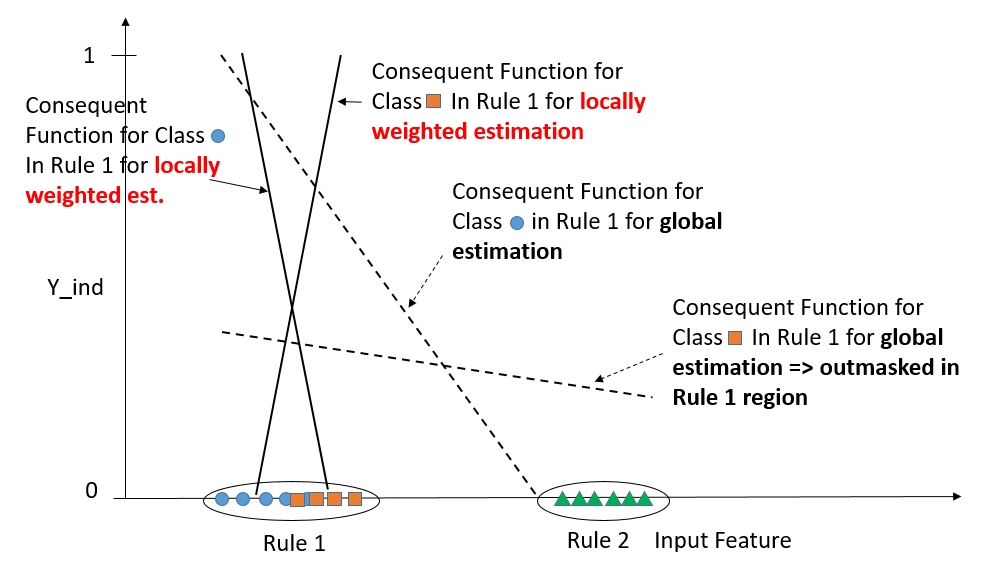}}
\caption{Consequent hyper-planes achieved when performing regression based on indicator entries subject to the circular and rectangular classes for Rule 1, one time with locally weighted rule activation levels (solid straight lines), one time with global (non-weighted) estimation (dashed straight lines).}
\label{local_vs_global_estimation_visexplain}
\end{figure}

The first term, i.e. the $\{0,1\}$ regression loss, is thus realized through a weighted least squares (WLS) functional, defined as:
\begin{equation}
\label{local_least_squares_problem_multilabel}
\min_{\vec{w}_{i,c}} \hspace{0.1cm} J_i(WLS) = \sum_{c=1}^K \sum_{k=1}^{N} \Psi_i(\vec{x}(k))(y_c(k)-f_{i,c}(\vec{w}_{i,c},\vec{x})(k))^2 \hspace{0.5cm} i=1,...,C
\end{equation}
with $C$ being the number of rules, $K$ the number of classes, $N$ the number of samples and $f_{i,c}(\vec{w}_{i,c},\vec{x})(k)$ the estimated value of the $k$th sample for the $c$th class based on the hyper-plane of the $i$th rule, and $y_c(k)$ the real (given) indicator value. This is only valid in the local region around the $i$th rule, therefore the rule activation level $\Psi_i(\vec{x}(k))$ is included as a sample weight in the WLS function: samples lying closer to the $i$th rule are considered much more during optimization than samples further away.
No interaction between the classes occurs here (this is shifted to the correlation-based learning term), and as the term within the outer sum is always positive due to the parabolic nature and because $\Psi_i(\vec{x}(k))$ is always greater than (or equal to) 0, the WLS objective function in \eqref{local_least_squares_problem_multilabel} can be optimized on each summand, i.e. on each class, separately.
Furthermore, we want to highlight that with locally weighted estimation, more flexibility in the consequent functions is achieved in the case of overlapping classes within a single rule, abandoning a possible out-masking effect as would be the case when using (non-weighted) global learning.
A two-dimensional visual explanation for this is shown in Figure \ref{local_vs_global_estimation_visexplain}.
Obviously, the rectangular class falling into Rule 1 is completely out-masked when applying global learning as its consequent function (dashed straight line) occurs completely below one of the circular class within the corresponding region. This is because the (green) samples belonging to the triangular class are used with full weights in the global estimation. In the case of locally weighted regression, these samples are down-weighed (close to 0), thus they do not influence the estimation, which is then more or less solely based on the samples from the circular and rectangular classes. This avoids the out-masking effect and induces greater flexibility to represent class overlaps within a single rule properly: obviously, the circular class is triggered when a sample falls into the left region of Rule 1 (as its consequent function delivers a higher value) and to the rectangular class when it falls into the right region of Rule 1, which is desired.

The second term is motivated by Lasso-based regularization, which is able to out-sparse as many parameters as possible. Thus, it yields sparse solutions with some interpretation capabilities to explain which input features for which classes in the corresponding rules are important --- please note that among the local regions, the feature importance on the classes may vary, which in turn motivates the use of a localized Lasso-based regularizer, which (for the $i$-th rule) is given by:
\begin{equation}
\label{lasso_regularization}
J_i(REG) = \alpha \sum_{c=1}^K \sum_{j=1}^p  |w_{ij,c}|
\end{equation}
with $\alpha$ being a tunable regularization parameter.

The third term is motivated by the field of correlation-based learning in multi-label classification settings to emphasize the possible interrelations between class labels. This is a peculiar characteristic of multi-label classification problems \cite{ZhangYeung13} and typically increases the accuracy of the classifiers, as shown in \cite{JiTangYuYe08} \cite{LiWangPavluAslam16}.
Based on the analysis in \cite{ZhangYeung13}, we exploit the assumption that the correlation between two labels should be consistent with the correlation between their discriminative features. In our context of consequent learning, this means that the consequent vectors for two classes should have a high overlap, and thus should be similar, when the class labels are highly correlated. This is simply because the linear combination of the input features as outputs of the consequent functions $f_{i,c}(\vec{w}_{i,c},\vec{x})$ should lead to similar prediction results for the two classes. As we perform local learning per rule, weighted correlation has to be considered (again with weights $\Psi_i(\vec{x})$ as in WLS, see above) and thus the objective term for the $i$th rule is constructed as follows:
\begin{equation}
\label{correlation_based_objterm}
J_i(CORR) = \sum_{c=1}^K \sum_{d=1}^K a_{cd}(\Psi_i) <\vec{w}_{i,c},\vec{w}_{i,d}>
\end{equation}
with $<.,.>$ the scalar product between two vectors and $a_{cd}(\Psi_i) = 1-corr_{cd}(\Psi_i)$, where $corr_{cd}(\Psi_i)$ denotes the weighted correlation coefficient, which is calculated by:
\begin{equation}
\label{correlation_coefficient_empirical_weighted}
corr_{cd}(\Psi_i) = \frac{\sum_{k=1}^{N}\Psi_i(\vec{x}(k))(y_{c}(k)-\bar{y}_c)(y_d(k)-\bar{y}_d)}
{\sqrt{\sum_{k=1}^{N}\Psi_i(\vec{x}(k))(y_{c}(k)-\bar{y}_c)^2}\sqrt{\sum_{k=1}^{N} \Psi_i(\vec{x}(k))(y_d(k)-\bar{y}_d)^2}}
\end{equation}
with $\bar{y}_c$ being the weighted mean value of $y_c$, i.e. of the indicator vector for class $c$ on all samples, $\bar{y}_d$ the weighted mean value of $y_d$, i.e. of the indicator vector for class $d$ on all samples.
This means that when the weighted (local) correlation between classes $c$ and $d$ is low, $a_{cd}(\Psi_i)$ becomes high, thus $<\vec{w}_{i,c},\vec{w}_{i,d}>$ should be low in order to minimize \eqref{correlation_based_objterm}. However, this means that $\vec{w}_{i,c}$ and $\vec{w}_{i,d}$ should be dissimilar (90 degree angle between them), because then their scalar product is close to 0.

Combining the three objective terms together to one functional and exploiting a more compact formulation in matrix form, this leads to the following
objective function for the $i$th rule:
\begin{equation}
\label{objective_functional_overall}
\min_{W_{i}} \hspace{0.1cm} J_i = \frac{1}{2} \| (Y - W_i^TR^T) \sqrt{Q_i} \|^2_{F} + \alpha \|W_i\|_{1} + \beta Trace(A_iW_i^T W_i)
\end{equation}
with $W_i\in \mathbb{R}^{(p+1)\times K}$ being the consequent vectors for the $i$th rule stored column-wise per class, $Y\in \mathbb{R}^{K\times N}$ a matrix containing all indicator vectors for all classes as row vectors, $R\in \mathbb{R}^{N\times (p+1)}$ the regression matrix with the last column a vector containing values of ones (for the intercept) and the samples in the rows, $A_i\in \mathbb{R}^{K\times K}$ a matrix containing all $a_{cd}(\Psi_i)$'s for all class pairs and $Q_i\in \mathbb{R}^{N\times N}$ a weighting diagonal matrix defined by:
\begin{equation}
\label{weighting_matrix_locallearning}
Q_i = diag(\Psi_i(\vec{x}(1)), \Psi_i(\vec{x}(2)), ..., \Psi_i(\vec{x}(N)))
\end{equation}
$\sqrt{Q_i}$ denotes the square root of each diagonal element.

It can be shown that the second derivative of the first and the last term ($J_i(WLS)$ and $J_i(CORR)$) leads to a positive definite matrix $H = R^TQ_iR + \beta A_i$, due to $R^TQ_iR$ being positive definite as all entries in $Q_i$ are greater than 0 (due to the infinite support of multivariate Gaussians defined in \eqref{multivariate_Gaussian_distribution}) and $A_i$ a symmetric matrix with entries in $[0,2]$ --- the addition of $R^TQ_iR$ and $\beta A_i$ leads to a matrix pencil, whose positive definiteness is granted whenever $\beta > -1/\delta$ with $\delta$ being the minimum positive eigenvalue of the generalized eigenvalue problem $\lambda R^TQ_iR \vec{v} = A_i \vec{v}$ \cite{GolubVanLoan96}; this is the case in our setting, because $\beta$ is always greater than 0 to embed (local) correlation-based learning.
Therefore, the function $J_i(WLS) + J_i(CORR)$ is convex, and due to the fact that i) the $L_1$ norm is convex and ii) a linear combination of two convex functions is again convex according to analysis made in \cite{Zhang04}, the objective function in \eqref{objective_functional_overall} is convex.

\subsubsection{Batch Off-line Algorithm}
According to \eqref{objective_functional_overall}, it is easy to see that the first order derivative of $J(W_i^{(t)}) = J_i(WLS) + J_i(CORR)$ can be calculated by the following formula:
\begin{equation}
\label{derivative_J}
\bigtriangledown J(W_i^{(t)}) = R^TQ_iRW_i^{(t)} - R^TQ_iY^T + \beta W_i^{(t)}A_i
\end{equation}
Based on the considerations about Lipschitz continuity, which is the case when the first order derivative is bounded (proof is left to the reader), there exists a Lipschitz constant $Lip$, such that the second order Taylor expansion of $J = J_i(WLS) + J_i(CORR)$ can be represented as:
\begin{align}
\label{taylor_series_expansion}
\hat{J}(W_i) &= J(W_i^{(t)}) + <\bigtriangledown J(W_i^{(t)}), W_i - W_i^{(t)}> + \frac{Lip}{2} \| W_i - W_i^{(t)} \|^2_F   \nonumber \\
&= \frac{Lip}{2}  \| W_i - (W_i^{(t)} - \frac{1}{Lip} \bigtriangledown J(W_i^{(t)})) \|^2_F + const
\end{align}
with $W_i^{(t)}$ being the fixed point in the $t$th iteration.
The Lipschitz constant represents the second derivative of $J$ in $W_i^{(t)}$ and can be achieved by the maximal eigenvalues of the second derivative:
\begin{equation}
\label{Lipschitz_constant}
Lip = \sqrt{a_{max}(R^TQ_iR) + a_{max}(\beta A_i)}
\end{equation}
with $a_{max}(X)$ being the largest eigenvalue of matrix $X$.
Due to the Taylor series expansion in \eqref{taylor_series_expansion}, the solution in the $t+1$st iteration can be represented as:
\begin{align}
\label{taylor_series_expansion_iteration}
W_i^{(t+1)} &= \min_{W_{i}} \frac{Lip}{2}  \| W_i - (W_i^{(t)} - \frac{1}{Lip} \bigtriangledown J_i(W_i^{(t)})) \|^2_F + \alpha \|W_i\|_{1} \nonumber \\
&= \min_{W_{i}} \frac{1}{2} \| W_i - X_i^{(t)} \|^2_F + \frac{\alpha}{Lip} \|W_i\|_{1}
\end{align}
where $X_i^{(t)} = W_i^{(t)} - \frac{1}{Lip}\bigtriangledown J(W_i^{(t)})$.

Based on the use of the proximal gradient descent algorithm (as the $L_1$ norm is non-smooth) \cite{EfronHastieJohnstoneTibshirani04}, \eqref{taylor_series_expansion_iteration} can be solved by the following iterative thresholding algorithm to emphasize the sparsity of the solutions:
\begin{align}
X_i^{(t)} &= W_i^{(t)} - \frac{1}{Lip}\bigtriangledown J(W_i^{(t)}) \nonumber \\
W_i^{(t+1)} &= S_{\frac{\alpha}{Lip}} (X_i^{(t)})
\end{align}
where $S$ is the soft threshold function to shrink as many consequent parameters towards 0 as possible:
\begin{equation}
\label{soft_threshold_function}
S_{\frac{\alpha}{Lip}} (X_i^{(t)})_{ij}  =
\begin{cases} x_{ij} - \frac{\alpha}{Lip} & x_{ij} > \frac{\alpha}{Lip} \\
              x_{ij} + \frac{\alpha}{Lip} & x_{ij} < -\frac{\alpha}{Lip} \\
              0 & otherwise
\end{cases}
\end{equation}

Algorithm \ref{local_correlationbased_learning} summarizes the essential steps of the batch learning algorithm for multi-label correlation-based learning of the consequent parameters. The while-loop runs until convergence of the parameters, which, for instance, can be checked whenever the difference of the previous solution $W_i^{(t-1)}$ to the current solution $W_i^{(t)}$ is smaller than a small positive value: $\| W_i^{(t-1)} - W_i^{(t)} \| < \epsilon$.
The start value for the parameters, i.e. $W_i^{(0)}$, is established by solving the classical weighted least squares problem $J_i(WLS)$, thus by:
\begin{equation}
\label{WLS_solution_start}
W_i^{(0)} = (R^TQ_iR + \gamma I)^{-1} R^TQ_i Y
\end{equation}
where $\gamma$ is a regularization parameter to ensure a stable inverse of the matrix in the case of rank deficiencies; its concrete value is automatically achieved based on the condition of the matrix $R^TQ_iR$ as proposed and successfully evaluated in \cite{LughoferKindermann09jour}.
\eqref{WLS_solution_start} ensures a solution which has a closer relation to the final solution (than an arbitrary setting) and thus the convergence speed should be improved.
\begin{algorithm}[h!]
		\caption{Local Correlation-based Learning of Consequent Parameters (for the $i$th rule)}
		\label{local_correlationbased_learning}
		\begin{algorithmic}
			\STATE {\bfseries Input:} Input data matrix $R$, multi-label indicator matrix $Y$, weight matrix $Q_i$ as in \eqref{weighting_matrix_locallearning}, weighted anti-correlation matrix $A_i$ between class labels, parameters $\alpha$, $\beta$, current rule index $i$.
			\STATE {\bfseries Output:} Consequent parameter matrix $W_i$.
			\\~\\
			\STATE $W_i^{(0)} = (R^TQ_iR + \gamma I)^{-1} R^TQ_i Y$, with $\gamma$ calculated from the eigenvalues of $R^TQ_iR$ as in \cite{LughoferKindermann09jour}.
            \STATE Calculate the Lipschitz constant $Lip$ using \eqref{Lipschitz_constant}.
            \STATE $t = 0$.
			\STATE {\bf While not converged Do}
            \STATE \hspace{0.2cm} Calculate the derivative $\bigtriangledown J(W_i^{(t)})$ through \eqref{derivative_J}.
			\STATE \hspace{0.2cm} $X_i^{(t)} = W_i^{(t)} - \frac{1}{Lip}\bigtriangledown J(W_i^{(t)})$.
            \STATE \hspace{0.2cm} $W_i^{(t+1)} = S_{\frac{\alpha}{Lip}} (X_i^{(t)})$ using \eqref{soft_threshold_function}.
            \STATE \hspace{0.2cm} $t=t+1$.
            \STATE {\bf End}
		\end{algorithmic}
	\end{algorithm}

\subsubsection{Incremental Online Algorithm}
\label{incremental_corrbased_learning}
The online adaptation phase of consequent parameters is based on a single-pass and sample-wise update algorithm in order to meet the fast processing spirit of data streams: the single-pass nature ensures that no past samples and thus no (multiple) iterations over past samples are needed for parameter updates; the sample-wise nature ensures that each single sample can be processed through the algorithm without requiring any bigger data chunks.

The online update algorithm requires the update of three essential components: the update of the Lipschitz constant $Lip$ (as new samples may change the nature of the weighted Hessian and the weighted anti-correlation matrices), the update of the weighted anti-correlation matrix $A_i$, and the update of the components in $\bigtriangledown J(W_i)$ (which induces the update of the Hessian matrix $R^TQ_iR$ and the update of the $R^TQ_iY$ information matrix).

The update of the weighted anti-correlation matrix goes hand in hand with the update of the weighted correlations between each label pair, which in batch mode are calculated by \eqref{correlation_coefficient_empirical_weighted} (as $(A_i)_{cd} = 1 - corr_{cd}(\Psi_i)$. This includes one weighted covariance term (in the numerator) and two weighted variance terms (in the denominator).
Assuming $S_{\Psi}(N) = \sum_{k=1}^{N}\Psi_i(\vec{x}(k))$, the weighted variance over a class label $y_c$ can be reformulated in the following way:
\begin{align}
\label{weighted_variance_compactversion}
var_{c}(\Psi_i)(N) &= \sum_{k=1}^{N}\Psi_i(\vec{x}(k))(y_{c}(k)-\bar{y}_c(N))^2  \nonumber \\
&= \sum_{k=1}^{N}\Psi_i(\vec{x}(k))y_{c}^2(k) - 2\bar{y}_c(N)\sum_{k=1}^{N}\Psi_i(\vec{x}(k))y_c(k) + \bar{y}_c(N)^2 \sum_{k=1}^{N}\Psi_i(\vec{x}(k)) \nonumber \\
&= \sum_{k=1}^{N}\Psi_i(\vec{x}(k))y_{c}^2(k) - 2\bar{y}_c(N) S_{\Psi}(N)\bar{y}_c + \bar{y}_c^2(N) S_{\Psi}(N) \nonumber \\
&= \sum_{k=1}^{N}\Psi_i(\vec{x}(k))y_{c}^2(k) - \bar{y}_c(N)^2 S_{\Psi}(N)
\end{align}
where $\bar{y}_c(N)$ is the weighted mean $\bar{y}_c(N)=\frac{\sum_{k=1}^{N}\Psi_i(\vec{x}(k))y_{c}(k)}{\sum_{k=1}^{N}\Psi_i(\vec{x}(k))}$, which can be updated easily by:
\begin{equation}
\label{incremental_weighted_mean}
\bar{y}_c(N) = \frac{\bar{y}_c(N-1)*S_{\Psi}(N-1) + \Psi_i(\vec{x}(N))*y_{c}(N)}{S_{\Psi}(N)}
\end{equation}
Then, the following derivation can be made for updating the weighted variance term in the denominator of \eqref{correlation_coefficient_empirical_weighted}:
\begin{align}
\label{incremental_weighted_variance}
var_{c}(\Psi_i)(N) - var_{c}(\Psi_i)(N-1) &= \sum_{k=1}^{N}\Psi_i(\vec{x}(k))y_{c}^2(k) - S_{\Psi}(N) \bar{y}_c^2(N) - \sum_{k=1}^{N-1}\Psi_i(\vec{x}(k))y_{c}^2(k) + S_{\Psi}(N-1) \bar{y}_c^2(N-1) \nonumber \\
&= S_{\Psi}(N)y_{c}^2(N) - S_{\Psi}(N) \bar{y}_c^2(N) + S_{\Psi}(N-1) \bar{y}_c^2(N-1) \nonumber \\
&= S_{\Psi}(N)y_{c}^2(N) - S_{\Psi}(N) \bar{y}_c^2(N) + (S_{\Psi}(N)-\Psi_i(\vec{x}(N)) \bar{y}_c^2(N-1) \nonumber \\
&= \Psi_i(\vec{x}(N))(y_{c}^2(N) - \bar{y}_c^2(N-1)) + S_{\Psi}(N)(\bar{y}_c^2(N-1) - \bar{y}_c^2(N)) \nonumber \\
&= \Psi_i(\vec{x}(N))(y_{c}^2(N) - \bar{y}_c^2(N-1)) + S_{\Psi}(N)(\bar{y}_c(N-1) - \bar{y}_c(N))(\bar{y}_c(N-1) + \bar{y}_c(N)) \nonumber \\
&=^* \Psi_i(\vec{x}(N))(y_{c}^2(N) - \bar{y}_c^2(N-1)) + \Psi_i(\vec{x}(N))(\bar{y}_c(N-1) - y_c(N))(\bar{y}_c(N-1) + \bar{y}_c(N)) \nonumber \\
&= \Psi_i(\vec{x}(N))(y_c(N) - \bar{y}_c(N-1))(y_c(N) - \bar{y}_c(N)) \nonumber \\
& \Rightarrow \nonumber \\
var_{c}(\Psi_i)(N) &= var_{c}(\Psi_i)(N-1) + \Psi_i(\vec{x}(N))(y_c(N) - \bar{y}_c(N-1))(y_c(N) - \bar{y}_c(N))
\end{align}
This finally means that the variance is updated by adding the weighted variance contribution of the new sample to the old variance, whereas this contribution importantly contains the {\bf old} weighted mean $\bar{y}_c(N-1)$ in one part and the new (updated) weighted mean in the other part.
Please note that in the case of intending to achieve the real weighted variance, $var_{c}(\Psi_i)(N)$ can be divided through $S_{\Psi}(N)$ in each iteration.
The *-sign in \eqref{incremental_weighted_variance} indicates a (probably) not obvious equivalence that $S_{\Psi}(N)(\bar{y}_c(N-1) - \bar{y}_c(N)) =
\Psi_i(\vec{x}(N)) (\bar{y}_c(N-1) - y_c(N))$. This can be shown by:
\begin{align}
\label{incremental_weighted_var_addon}
S_{\Psi}(N)\bar{y}_c(N-1) - S_{\Psi}(N)\bar{y}_c(N) &=  S_{\Psi}(N)\bar{y}_c(N-1) - \sum_{k=1}^{N}\Psi_i(\vec{x}(k))y_{c}(k) \nonumber \\
& = \frac{(S_{\Psi}(N-1) + \Psi_i(\vec{x}(N))) \sum_{k=1}^{N-1}\Psi_i(\vec{x}(k))y_{c}(k)}{S_{\Psi}(N-1)} - \nonumber \\
& \hspace{0.4cm} \frac{S_{\Psi}(N-1) \sum_{k=1}^{N-1}\Psi_i(\vec{x}(k))y_{c}(k) - S_{\Psi}(N-1) \Psi_i(\vec{x}(N))y_{c}(N)}{S_{\Psi}(N-1)} \nonumber \\
&= \frac{\Psi_i(\vec{x}(N)) \sum_{k=1}^{N-1}\Psi_i(\vec{x}(k))y_{c}(k) - S_{\Psi}(N-1) \Psi_i(\vec{x}(N))y_{c}(N)}{S_{\Psi}(N-1)} \nonumber \\
&= \frac{\Psi_i(\vec{x}(N)) \left(\sum_{k=1}^{N-1} \Psi_i(\vec{x}(k))y_{c}(k) - S_{\Psi}(N-1)y_{c}(N)\right)}{S_{\Psi}(N-1)} \nonumber \\
&= \Psi_i(\vec{x}(N))(\bar{y}_c(N-1) - y_c(N))
\end{align}

The incremental weighted covariance-like update formula (for the numerator in \eqref{correlation_coefficient_empirical_weighted}, without division through $S_{\Psi}(N)$) can be derived in the same manner as the weighted variance (again with the covariance contribution in one term using the old weighted mean) and is given by:
\begin{equation}
\label{incremental_weighted_covariance}
covar_{c,d}(\Psi_i)(N) = covar_{c,d}(\Psi_i)(N-1) + \Psi_i(\vec{x}(N))(y_c(N) - \bar{y}_c(N-1))(y_d(N) - \bar{y}_d(N))
\end{equation}

Regarding the update of the Hessian matrix $R^TQ_iR$ (required for updating the derivative $\bigtriangledown J(W_i)$ and the Lipschitz constant), this can be accomplished by the same formula as for the weighted covariance matrix using a weighted mean of 0 instead of the real weighted mean of the features. An alternative is to update the inverse Hessian matrix as used in recursive fuzzily weighted least squares (see below) and to invert this matrix after each update step. This, however requires cubic computing time $O((p+1)^3)$ instead of a quadratic one, but depending solely on $p$ due to the local learning effect.
The update of the information matrix $R^TQ_iY^T$ as used in the calculation of $\bigtriangledown J(W_i)$ (second term), is obviously achieved by:
\begin{equation}
\label{information_matrix_update}
R^TQ_iY^T(N) = R^TQ_iY^T(N-1) +  \vec{r}(N)^T\Psi_i(\vec{x}(N))\vec{y}(N)
\end{equation}
with $\vec{r}(N) = [x_1 \hspace{0.1cm} ... \hspace{0.1cm} x_p \hspace{0.1cm} 1]$ the regressor row vector in the current sample (the last entry for the intercept) and $\vec{y}(N)$ the multi-label indicator row vector in the current sample.

Furthermore, according to the same consideration as for the initial batch offline phase, we always perform a pure weighted least-squares (WLS) based update step on a new sample before applying the correlation-based update to achieve updated consequent parameters.
The WLS-based update step is achieved in a single-pass convergent manner by employing the recursive fuzzily weighted least squares (RFWLS) estimator \cite{AngelovFilev04} (used in most of the current evolving neuro-fuzzy systems approaches (see \cite{LughoferChapter14})):
\begin{equation}
\label{recursive_weighted_param_update}
W_i(N) = W_i(N-1)+\gamma(N-1)(\vec{y}(N)-\vec{r}(N)W_i(N-1))
\end{equation}
\begin{equation}
\label{recursive_weighted_gamma_vector}
\gamma (N-1) = \frac{P_i(N-1)\vec{r}(N)}{\frac{1}
{\Psi_i(\vec{x}(N))}+\vec{r}^T(N)P_i(N-1)\vec{r}(N)}
\end{equation}
\begin{equation}
\label{recursive_weighted_inv_hesse_update}
P_i(N) = (I-\gamma (N-1)\vec{r}^T(N))P_i(N-1)
\end{equation}
with $P_i(N)=(R(N)^TQ_i(N)R(N))^{-1}$ the inverse weighted Hessian matrix, directly updated through the last formula (no matrix inversion needed).
Please note that here the conventional RFWLS formulas operating in vector form are simply extended to a matrix form, where one column of $W_i(N-1)$ represents the consequent vector for the corresponding class and where one column of the row vector $(\vec{y}(N)-\vec{r}(N)W_i(N-1))$ represents the prediction error in the current sample for the corresponding class. This is multiplied by a column vector $\gamma(N-1)$ denoting the Kalman gain and used as a multiplication factor to steer the intensity of the parameter update.

Algorithm \ref{local_correlationbased_learning_incremental} summarizes the essential steps of the incremental correlation-based update of the consequent parameters in a single rule (in the $i$th one); this algorithm can be performed for all $C$ rules independently due to iteration from $i=1$ to $i=C$.
\begin{algorithm}[h!]
		\caption{Incremental Local Correlation-based (ILC) Learning of Consequent Parameters (for the $i$th rule)}
		\label{local_correlationbased_learning_incremental}
		\begin{algorithmic}
			\STATE {\bfseries Input:} Input data sample $(\vec{x},\vec{y})$, regressor $\vec{r} = [\vec{x}\hspace{0.2cm} 1]$, weighted anti-correlation matrix $A_i(N-1)$ between class labels, current consequent parameter matrix $W_i(N-1)$, current information matrix $R^TQ_iY(N-1)$ and (inverse) Hessian matrix $P_i(N-1)$, parameters $\alpha$, $\beta$, current rule index $i$.
			\STATE {\bfseries Output:} Updated consequent parameter matrix $W_i(N)$ and updated help matrices.
			\\~\\
            \STATE Calculate normalized rule activation level $\Psi_i(\vec{x})$ according to \eqref{multilabel_classifiers_inference}.
            \STATE Perform a recursive fuzzily weighted least squares step by applying \eqref{recursive_weighted_param_update} to \eqref{recursive_weighted_inv_hesse_update} $\rightarrow$ achieving $W_i(N)^{(0)}$.
            \STATE Update the weighted variance term of all classes using \eqref{incremental_weighted_variance}.
            \STATE Update the weighted co-variance term of all class pairs using \eqref{incremental_weighted_covariance}.
            \STATE Update the weighted anti-correlation matrix $A_i(N-1)$ using weighted correlation coefficients as in \eqref{correlation_coefficient_empirical_weighted} $\rightarrow$ $A_i(N)$.
            \STATE Update $R^TQ_iR$ using \eqref{incremental_weighted_variance} and \eqref{incremental_weighted_covariance} with zero means on the regressors; or alternatively set $R^TQ_iR = P_i(N)^{-1}$ (with $P_i(N)$ updated in the second step above) as this matrix is positive-definite and thus its inversion is stable.
            \STATE Recalculate the Lipschitz constant $Lip$ using \eqref{Lipschitz_constant}.
            \STATE Update the information matrix $R^TQ_iY(N-1)$ using \eqref{information_matrix_update} $\rightarrow$ $R^TQ_iY(N)$.
            \STATE $t = 0$.
			\STATE {\bf While not converged Do}
            \STATE \hspace{0.2cm} Calculate the derivative $\bigtriangledown J(W_i(N)^{(t)})$ through \eqref{derivative_J}.
			\STATE \hspace{0.2cm} $X_i^{(t)} = W_i(N)^{(t)} - \frac{1}{Lip}\bigtriangledown J(W_i(N)^{(t)})$.
            \STATE \hspace{0.2cm} $W_i(N)^{(t+1)} = S_{\frac{\alpha}{Lip}} (X_i^{(t)})$ using \eqref{soft_threshold_function}.
            \STATE \hspace{0.2cm} $t=t+1$.
            \STATE {\bf End}
		\end{algorithmic}
	\end{algorithm}
This algorithm is carried out for each rule locally on each new incoming sample $(\vec{x},\vec{y})$. Thereby, the normalized rule activation level $\Psi_i(\vec{x})$ plays an essential role: for samples which are much further away from the $i$th rule than to other rules, $\Psi_i(\vec{x})$ approaches 0 and thus the intensity of the update of the consequent parameters of the $i$th rule automatically becomes tiny, which is desirable. According to our experience from several test experiments, it seems that usually one single step of the correlation-based update (last three steps in Algorithm \ref{local_correlationbased_learning_incremental}) is sufficient and stable.

An additional issue arises when a new rule is evolved by the evolving antecedent algorithm (see Section \ref{antecedent_learning}). In this case, the consequent parameter vectors for the new rule are set to those of the nearest rule, elicited through the Mahalanobis distance:
\begin{equation}
\label{consequent_parameters_initialset}
W_{C+1} = W_{win} \hspace{0.5cm} win = argmin_{i=1,...,C} \left((\vec{c}_{C+1}-\vec{c}_i)^T\Sigma_{i}^{-1}(\vec{c}_{C+1}-\vec{c}_i)\right)
\end{equation}
with $C$ the number of current rules and $\vec{c}_{C+1}$ the center of the new rule (typically set to the current data sample $\vec{x}$ upon rule evolution). This setting is in line with several past studies in the evolving fuzzy systems community to ensure robustness and quick convergence for new data samples (see \cite{LughoferChapter14}). In such a case, the incremental update based on Algorithm \ref{local_correlationbased_learning_incremental} continues with the next sample.

A further issue arises when two rules $i$ and $k$ are merged together by the evolving antecedent algorithm (see Section \ref{antecedent_learning}), due to cluster fusioning effects (see \cite{LughoferChapter14} \cite{SkrjancIglesiasLughoferGomide19} for an explanation of this particular occurrence and proper merging strategies).
In this case, the consequent vectors are merged by the strategy proposed in \cite{LughoferPositionPaper13}, which adequately resolves contradictory consequent functions (which is the case when their similarity is lower than the similarity of the antecedent parts); in consequent matrix notation, this is achieved by:
\begin{equation}
\label{rule_consequ_merging_mod}
W_{new} = W_i + \alpha \cdot \rho(W_i,W_k) \cdot (W_k - W_i),
\end{equation}
where $\alpha = k_k/(k_i + k_k)$ with $k_i$ being the support of the $i$th rule and $\rho(W_i,W_k)$ is a measure of consistency of the two rule consequents, typically calculated by comparing their similarity degree (measured through the dihedral angle) with the similarity of the rule antecedents (see \cite{LughoferPositionPaper13} for details).
For $\rho = 0$, indicating an inconsistency in the rule base, we obtain $W_{new} = W_i$, i.e., the consequent of the more relevant rule. For $\rho = 1$, on the other hand, \eqref{rule_consequ_merging_mod} reduces to the classical weighted average of both consequent parameters.


\section{Online Active Learning for Sample Selection}
\label{online_activelearning}
In (online) data stream classification problems, the labels may not necessarily be available automatically for each instance. Mostly, a data stream appears to be scarcely labelled or fully unsupervised because labeling the data requires manual intervention with significant and timely effort from experts or operators \cite{LughoferRichterNeisslHeidlEitzingerRadauer17}. This is much more severe than in the case of multi-label classification problems as often a few dozen labels have to be provided for a single sample \cite{WuShengZhangLi20}.
Online active learning is a methodology which weakens this problematic issue, as it selects those samples which are expected to help the classifier most to improve or at least maintain its predictive performance over time \cite{Lughofer17}. Depending on the percentage of the selected samples, the annotation effort can be more or less decreased. Typically, there exists a so-called {\em budget} in the form of an upper allowed percentage bound of samples to be annotated by the expert(s), which should not be exceeded due to her/his (their) interaction time constraints \cite{KremplKottkeLemaire15}.

Our novel approach handles online sample selection for fully unsupervised stream samples in combination with the architecture of multi-label fuzzy classifiers (to the best of our knowledge this has not been performed before) and is grounded on the following four basic concepts:
\begin{itemize}
\item Novelty content in the input space: this means that a sample represents a new state/condition and thus appears in a region of the feature space which is not covered well by the fuzzy rules extracted so far. In this sense, it is of utmost importance to enlarge the classifier to expand its knowledge.
\item Uncertainty in the output space: this means that a sample falls in an overlap region of two or more classes within a rule (see left part in Figure \ref{local_vs_global_estimation_visexplain}) or in between rules and is thus close to the decision boundary and therefore produces ambiguous outputs.
\item Instability in the parameter space: this means that a sample falls into a sparsely populated region and thus may increase sample significance and thus parameter robustness in this region.
\item Budget-based selection: dynamically ensures that the budget pre-defined by a user is never exceeded.
\end{itemize}

\subsection{Novelty Content in the Input Space}
\label{novelty_content_inputspace}
\begin{figure}[t]
\centering
\includegraphics[width=0.45\textwidth]{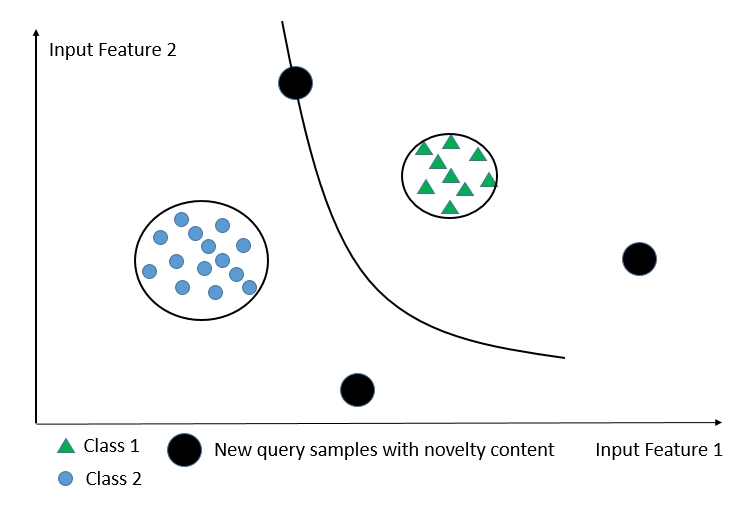}
\caption{Two-dimensional example of three new query samples (shown as big dots) denoting significant novelty content with respect to the rules' range of influence shown with solid ellipsoids; selecting such samples for class label annotation would help greatly to learn new rules properly and 'bend' the decision boundary into the correct direction.}
\label{novelty_content_explaination}
\end{figure}

The first criterion for online sample selection relies on the aspect that a sample which denotes a high novelty content for the current multi-label fuzzy classifier is a valuable one for expanding the (knowledge of the) classifier to a new region and finally to reduce possible extrapolation on further query samples. The latter is typically a precarious issue, as the higher the extrapolation degree, the lower the certainty in the output class and typically also the lower the classification accuracy. A two-dimensional example of this aspect is visualized in Figure \ref{novelty_content_explaination}, where new incoming samples (bigger dots) denote high novelty content subject to the available fuzzy rules (extracted before on older data samples) with their contours making their coverage spread shown as solid ellipsoids. Intuitively, the classification decision in these samples is very uncertain, although they may be safely located on one side of the decision boundary. Hence, it is different from classical uncertainty criteria, often based on the closeness to the decision boundary, as used in many active learning approaches (see \cite{Lughofer17}).

The decision as to whether a new sample denotes sufficient novelty content for the current fuzzy classifier
can be directly associated with the rule evolution condition in a natural way, which in our case is defined in \eqref{rule_evolution_criterion} (so, no additional criterion needs to be developed as would be the case for non-evolving, pure incremental classifiers). This is because this condition actually already checks for the novelty content within the embedded evolving learning algorithm and whenever it is met, a new rule is evolved. For such a new rule appearing in a new region, its consequent parameters should be properly learned, thus the class labels are required in the update through the use of Algorithm \ref{local_correlationbased_learning_incremental}.
Once a sample is selected based on its novelty content (thus meeting \eqref{rule_evolution_criterion}), all labels are required to be annotated by a user ({\em full annotation} of a sample).

\subsection{Uncertainty in the Output Space}
\label{uncertainty_outputspace}
\begin{figure}[t]
\centering
\includegraphics[width=0.4\textwidth]{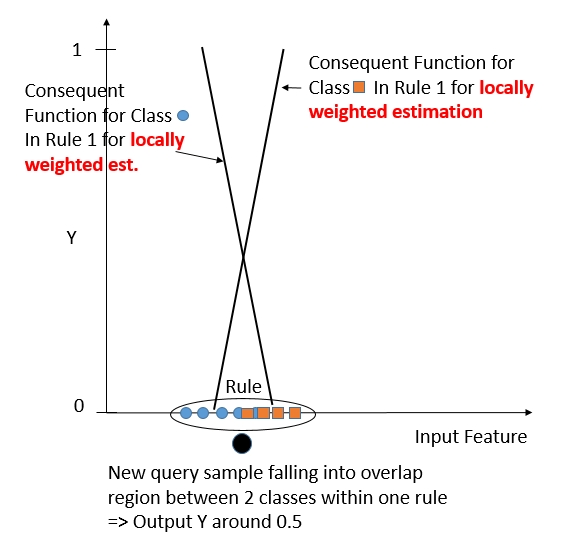}
\includegraphics[width=0.53\textwidth]{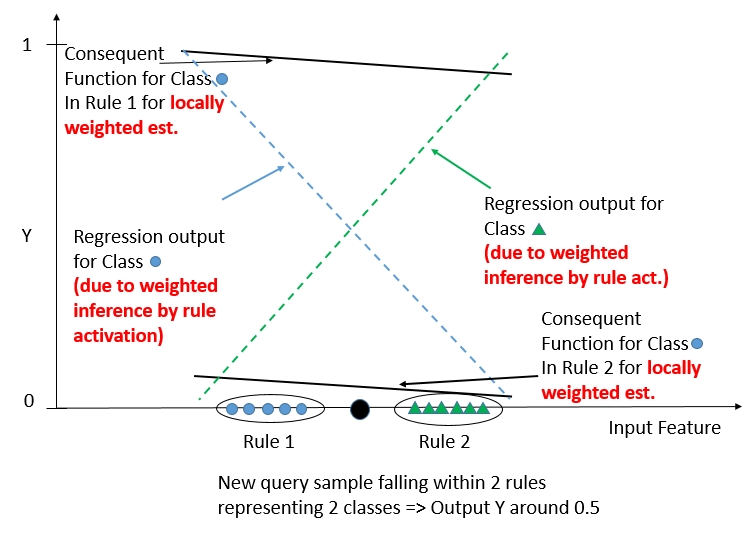}
\caption{Left: a case where a sample falls within an overlap region of a single rule and thus produces an ambiguous output (around 0.5 for both classes); right: a case where a samples falls within two rules representing two distinct classes and thus produces an ambiguous overall output (around 0.5 to both regression surfaces for the two classes shown as dashed lines), as it is calculated through the weighted inference (with weights equal to the rule activation levels) by \eqref{multilabel_classifiers_inference}}
\label{uncertainty_outputspace_explaination}
\end{figure}

The second criterion for online sample selection relies on the uncertainty in the output space when a new sample is classified by the current classifier. Such uncertainty can also arise in the case where there is no novelty content, but the sample falls into an overlap region of two or more classes or falls in between two rules (with non-significant novelty content) which represent two distinct classes (see Figure \ref{uncertainty_outputspace_explaination} for a two-dimensional example with new samples increasing output uncertainty shown as large dots).

In our case of multi-output regression through consequent hyper-planes, the output uncertainty can be checked through the ambiguity of continuous regression outputs $\hat{\vec{y}}$ as produced through the fuzzy inference in \eqref{multilabel_classifiers_inference}. According to \eqref{multilabel_output}, ambiguity is present when one or more continuous outputs are lying close to 0.5, i.e. close to the threshold for deciding 0 (not belonging to the corresponding class) or 1 (belonging to the corresponding class). A visual explanation becomes obvious from Figure \ref{uncertainty_outputspace_explaination}. Thus, the second selection criterion becomes:
\begin{equation}
\label{second_selection_criterion}
\exists_{j=1,...,K} \hspace{0.2cm} (\hat{\vec{y}}_j < thresh2) \wedge (\hat{\vec{y}}_j > 1-thresh2)
\end{equation}
Whenever one or a few $\hat{\vec{y}}_j$'s fulfill this criterion, it is an open question as to whether the annotation effort is restricted to the corresponding labels ({\em partial annotation} of a sample) or to the whole sample, i.e. whether all labels need to be checked by a user. This is possibly dependent on the concrete application scenario and associated multi-label characteristics.
Therefore, in the experiment results section, we foresee a labels-based and a samples-based evaluation, where $Thresh2$ was set to a fixed value of 0.6 in all experiments.

\subsection{Instability/Uncertainty in the Parameter Space}
\label{instability_paramspace}
The third criterion for online sample selection relies on the instability of the consequent parameters estimated by Algorithm \ref{local_correlationbased_learning_incremental} throughout the stream learning process. Such instability may arise due to sample sparsity and/or sufficient noise contained in the data. In particular, consider the case of increasing novelty content contained in new samples, based on which a new rule is evolved: at the beginning, there exists only one or a couple of samples representing the new knowledge/state; thus, the consequent parameters of the new rule are usually somewhat insignificant and uncertain.

To decrease this unpleasant occurrence efficiently, we aim to select samples which significantly reduce the uncertainty and thus the instability of the parameters {\em directly}. So, we do not use a sparsity criterion as occurs in many active learning approaches \cite{DonmezCarbonell10}, because sparsity is not a necessary condition to reduce parameter instability; for instance, a sample falling very close to a previously selected sample in a still sparse region will not further increase sample variety there, nor will it reduce (consequent) parameter uncertainty in this region; and furthermore, the distance may not be the only criterion for a sample to become very valuable for decreasing parameter instability in sparse regions. Moreover, already dense regions may also lead to instable parameter estimations, depending on their noise content and variety. Thus, a criterion for decreasing parameter uncertainty/instability {\em directly} is a much more general and efficient one.

It is well-known from the literature that the {\em Fisher information matrix} gives statistical evidence about the information content of the underlying data and the co-variance of the estimated parameters \cite{Frieden04}, thus it can be seen as a reliable measure for parameter uncertainty: high variances usually point to high parameter uncertainties. Deduced from the global Fisher information matrix \cite{Frieden04}, the local Fisher information matrix for the $i$th rule is defined by:
\begin{equation}
\label{Fisher_information_matrix_local}
F_i = \sum_{k=1}^N \Psi_i(k) \frac{\partial \hat{y}(\vec{x}_k)}{\partial \Phi} \frac{\partial \hat{y}(\vec{x}_k)}{\partial \Phi}^T
\end{equation}
with $\Phi$ being the vector of the parameters under study and $\Psi_i(k)$ being the normalized rule activation level.
In the case of {\em linear} consequent parameters, the local (rule-based) Fisher information matrix falls together with the Hessian matrix $R^TQ_iR$, whose inverse $P_i$ is updated through the RFWLS step \eqref{recursive_weighted_inv_hesse_update} on each sample by Algorithm \ref{local_correlationbased_learning_incremental}.

In a (batch) model-based design of experiments, where the input regions are also searched to identify from where to best select the samples, usually several optimality criteria acting on the Fisher information matrix are employed to steer the input control signal, such as A-optimality, E-optimality, D-optimality, to name a few. A-optimality aims to minimize the trace of the inverse of $F$, whereas most of the other criteria are operating on the original matrix $F$ (Hessian). Hence, to avoid matrix back-inversion after each update step of the inverse, we apply the A-optimality criterion to check whether a significant downtrend in the trace of $P_i$ can be observed. This is performed for each rule.
Thus, a sample is selected whenever
\begin{equation}
\label{downtrend_Fisher_criterion}
\max_{i=1,...,C} \left(\frac{trace(P_i(N-1))-trace(P_i(N))}{trace(P_i(N-1))}\right) > thresh3
\end{equation}
with $P_i(N)$ being the updated matrix on the new sample using \eqref{recursive_weighted_inv_hesse_update} (which is a fully unsupervised formula) and $thresh3$ a threshold denoting a significant percentual downtrend, typically lying between 5\% and 10\% (we used $thresh3 = 0.075$, thus 7.5\% as the threshold in all experiments).
As in the case of novelty content, all the corresponding class labels need to be assigned once a sample is selected.

\subsection{Budget-Based Selection}
\label{budgetbased_selection}
An important issue in active sample selection is the limitation of the number of selected samples, typically in the form of a maximal percentage the user has the ability to annotate (due to effort and time constraints). Such a limitation is also termed as {\em budget} in the active learning community \cite{ZhangZhaoNiu19}.

To meet a budget-based limitation for arbitrary data streams despite a reliable fixed setting of selection thresholds (as discussed above), we recalculate the labelling expense for each new incoming sample $\vec{x}(N)$ as $Sel(N)/N$, with $N$ being the number of samples seen so far and $Sel(N)$ the number of selected samples, and simply check whether the budget would be exhausted when the next sample is selected. Only if this is not the case, the sample is selected whenever it meets one of the three criteria discussed above, i.e. \eqref{rule_evolution_criterion}, \eqref{second_selection_criterion} and \eqref{downtrend_Fisher_criterion}.
However, such situations can jeopardize the classifier's significance and accuracy, because the training process may be capped off too early when the classifier is not yet mature enough to select samples properly \cite{PratamaLuLughoferZhangAnavatti16}. Therefore, we adopt the variable sampling strategy according to the suggestions in \cite{ZliobaiteBifetPfahringerHolmes14}, which aims to counterbalance the actual labelling expense and in turn consumes the budget more uniformly irrespective of the types of possible dynamics/changes contained in the data.
This is achieved by adapting the thresholds in \eqref{second_selection_criterion} and \eqref{downtrend_Fisher_criterion} autonomously based on the level of exhaustiveness of the budget (i.e. increasing them when the exhaustiveness becomes closer) and based on the recent selected samples.


\section{Experiment Setup}
\label{experimental_setup}
\subsection{Data Sets}
We used five high-dimensional multi-label data sets from the multi-label classification data set repository\footnote{http://mulan.sourceforge.net/datasets-mlc.html} to obtain a feeling about the algorithm's behavior etc. and to check for the performance of the classifier updates on a portion of separate test samples, serving as online stream samples when being sent sample-wise into the update of the multi-label fuzzy classifier.
These data sets comprise:
\begin{itemize}
\item Birds data set: This is a data set to predict the set of bird species that are present, given a ten-second audio clip; the multi-label aspect comes in as birds may belong to various species based on different properties.
\item Emotions data set: This is a data set to classify music into the emotions it evokes according to the Tellegen--Watson--Clark model of mood: amazed-surprised, happy-pleased, relaxing-calm, quiet-still, sad-lonely and angry-aggressive. It consists of 593 songs with 6 classes, where different emotions can be evoked.
\item Water quality data set: This is used to predict the quality of water of Slovenian rivers, knowing 16 characteristics such as the temperature, PH, hardness, NO2 or C02; the quality thus depends on multi(-label) factors.
\item Yeast data set: This contains micro-array expressions and phylogenetic profiles for 2417 yeast genes. Each gene is annotated with a subset of 14 functional categories (e.g., metabolism, energy, etc.) of the top level of the functional catalogue.
\item Mediamill: This is a large-label-scale multimedia data set for generic video indexing, which was extracted from the TRECVID 2005/2006 benchmark. This data set contains 85 hours of international broadcast news data categorized into 100 labels and each video instance is represented as a 120-dimensional feature vector of numeric features.
\end{itemize}
The characteristics of the data sets are listed in Table \ref{summary_data_sets_characteristics}.
\begin{table}[t]
\begin{center}
\caption{{\small Characteristics of the applied data sets.}}
\label{summary_data_sets_characteristics}
{\small
\begin{tabular}{|l|c|c|c|}
\hline
Data Set & Input Dim. & \# of Samples & \# of Classes \\
\hline \hline
Birds & 260 & 645 & 19 \\
\hline
Emotions & 72 & 593  & 6  \\
\hline
Water Quality & 16 & 1060 & 14 \\
\hline
Yeast & 103 & 2417 & 14 \\
\hline
Mediamill & 120 & 43907 & 101\\
\hline
\end{tabular} }
\end{center}
\end{table}

\subsection{Evaluation Strategy and Measures}
\label{evaluation_measures}
The evaluation of our online multi-label classifier approach is grounded on the analysis of the performance trend lines over new incoming test samples which are used as pseudo-stream samples by sending them sample-wise into the evolving antecedent learning engine and incremental consequent parameter update algorithm listed in Algorithm \ref{local_correlationbased_learning_incremental}.
Two basic measures, widely used in the multi-label classification community \cite{ZhangZhou14}, serve as the basis for checking the classification performance of the multi-label classifier:
\begin{itemize}
\item Average precision (AP): This evaluates the ratio of the related label ranking before a certain label $l$:
    \begin{equation}
    \label{average_precision}
    AP = \frac{1}{N} \sum_{k=1}^N \frac{1}{|L_{\vec{x}(k)}|} \sum_{l \in L_{\vec{x}(k)}} \frac{|\{l^*\in L_{\vec{x}(k)} | \hat{y}_{l^*}(k) \geq \hat{y}_{l}(k)\}|}{rank_l(k)}
    \end{equation}
    where $N$ is the number of samples, $L_{\vec{x}(k)}$ is the set of labels associated with sample $\vec{x}(k)$, $L_{\vec{x}(k)} = \{y_c(k), c=1,...,K | y_c = 1\}$, $\hat{y}_{l^*}(k)$ is the predicted continuous output of the classifier in the $k$th sample to the $l^*$-th label and $rank_l(k)$ is the ranking of label $l$ in the $k$th sample according to a descending list of continuous (predicted) output values over all labels, $rank_l(k) = pos(sort(\hat{\vec{y}}(k)))$, with $pos$ the position in the sorted list of $\hat{\vec{y}}$'s.
    The larger the value of AP, the better the classification performance.
\item Partial accuracy (PA): This evaluates the ratio between correctly predicted classes in each sample compared to the overall number of classes and sums this up for all samples, thus is given by:
    \begin{equation}
    \label{partial_accuracy}
    PA =  \frac{1}{KN} \sum_{k=1}^N |\{\mathbb{I}_{\hat{\vec{y}}_c(k) == \vec{y}(k)} = 1\}|
    \end{equation}
    with $K$ being the number of classes, $N$ the number of samples and $\mathbb{I}$ the indicator function on vectors, delivering 1 when the condition in its subscript is fulfilled, 0 otherwise; '==' denotes the element-wise operation on the two vectors (comparing all the elements in the same positions, i.e. corresponding to the same labels).
    The partial accuracy therefore checks how much overlap between the observed and the predicted classes is achieved and thus may enjoy greater intuitive explainability of the classifier accuracy than the average precision.
\end{itemize}

In a data streaming environment, the samples come in one by one, thus the accuracy of a classifier should be checked anytime based on the (number of) samples seen so far, yielding an accuracy trend line over time. This requires a permanent update of both measures.
We used the interleaved-test-and-then-train scenario, which is a widely established scenario in online learning environments. Thereby, for each incoming sample $\vec{x}$ first a prediction is made, the accuracy measure is updated based on the real observed value(s) in the incoming sample (included in all the data streams used for evaluation purposes) and then the classifier is updated with this sample, either always when not using any online active sample strategy or only upon selection according to the three AL criteria discussed in Section \ref{online_activelearning}.
In this sense, the update of the average precision with a new incoming sample $\vec{x}(N+1)$ is achieved by the following formula:
\begin{equation}
AP(N+1) = \frac{AP(N)*N + \frac{1}{|L_{\vec{x}(N+1)}|} \sum_{l \in L_{\vec{x}(N+1)}} \frac{|\{l^*\in L_{\vec{x}(N+1)} | \hat{y}_{l^*}(N+1) \geq \hat{y}_{l}(N+1)\}|}{rank_l(N+1)}}{N+1}
\end{equation}
and the update of the partial accuracy by:
\begin{equation}
\label{update_partial_accuracy}
PA(N+1) = \frac{PA(N)*KN + |\{\mathbb{I}_{\hat{\vec{y}}_c(N+1) == \vec{y}(N+1)} = 1\}|}{K(N+1)}
\end{equation}
where both are initialized to 0, i.e. $AP(0) = 0$ and $PA(0) = 0$.

We applied an initial batch of data for an initial training step, where the learning parameters $(\alpha,\beta,vigilance)$ are tuned, the latter steering the evolution of the rules (the lower, the more rules are evolved), within a grid-search based cross-validation phase. The learning parameter settings leading to the lowest cross-validation error are used for the online update and evolution phase on the remaining data sets. The grid points for the three parameters are comprehensively set in the following way: $\alpha=[0 \hspace{0.1cm} 0.01 \hspace{0.1cm} 0.025 \hspace{0.1cm} 0.05 \hspace{0.1cm} 0.075 \hspace{0.1cm} 0.1 \hspace{0.1cm} 0.5 \hspace{0.1cm} 1 \hspace{0.1cm} 5 \hspace{0.1cm} 10], \beta = [0 \hspace{0.1cm} 0.1 \hspace{0.1cm} 0.5 \hspace{0.1cm} 1 \hspace{0.1cm} 5 \hspace{0.1cm} 10 \hspace{0.1cm} 50 \hspace{0.1cm} 100], vigilance = [0.1 \hspace{0.1cm} 0.15 \hspace{0.1cm} 0.2 \hspace{0.1cm} 0.25 \hspace{0.1cm} ... \hspace{0.1cm} 0.9]$.
We compared our approach with the following related state-of-the-art multi-label classification variants:
\begin{itemize}
\item One-versus-rest classification is a widely used variant to resolve multi-class problems and as suggested for the online evolving case in \cite{AngelovLughoferZhou09}, for standard multi-class classification problems (no multi-labels are investigated, just a single class vector): therein, the consequent functions are estimated separately and independently for each class, which thus can also be directly applied to the multi-label based indicator vectors stored in $Y$ (column-wise estimation).
\item Classifier chaining is a widely used variant to resolve multi-label problems \cite{ReadPfahringerHolmes11}: it performs a direct consequent learning for the first class solely based on the input feature space, a consequent learning for the second class based on the feature space joint with the indicator vector of the first class, a consequent learning for the third class based on the feature space joint with the indicator vectors of the first and second class and so on; thus, it internally considers some relations among labels when learning for subsequent classes (due to joint feature spaces), however it does not integrate an explicit objective term (and associated optimization) for label correlation/relation learning.
\end{itemize}
For both variants, we apply the same rule evolution and antecedent learning engine as in our approach, which yields a direct comparison of how much improvement is achieved by the advanced incremental consequent learning through the inclusion of label correlation and Lasso-type shrinkage of the parameters (the main novelty of this paper together with the online active learning strategy for multi-label fuzzy classifiers). This is because for the related SoA variants, the classical RFWLS method is solely used for updating the consequent parameters for each class separately.
For both variants, we also perform an initial tuning phase of the essential parameters, which is left to the $vigilance$, using the same grid as in our approach.

Finally, we performed several tests with our online active sample selection approach, divided into a labels-based and a samples-based budget strategy: the labels-based budget restricts the number of labels selected so far to a maximal percentage subject to the number of labels seen so far (equal to $KN$), whereas the samples-based budget restricts the number of samples selected so far. The first strategy thus assumes that single labels can be annotated by a user for a particular sample by completely ignoring the other (non-selected) labels.
We applied various budgets: 10\%, 20\% and 30\% and compared their resulting partial accuracy trend lines with those achieved by a full classifier update on all samples (to see the extent of accuracy decrease with which effort), while reporting the actual selected samples (always smaller than the budget).
In the first run, we only used the first two criteria (due to the 'novelty content in the input space' and 'uncertainty in the output space'), whereas in an additional run, we demonstrate how much accuracy improvement can be achieved when using the advanced concept of uncertainty in the parameter space.

\section{Results}
\label{results}
\subsection{Permanent Model Updates}
\begin{figure}[t]
\centering
\includegraphics[width=0.46\textwidth]{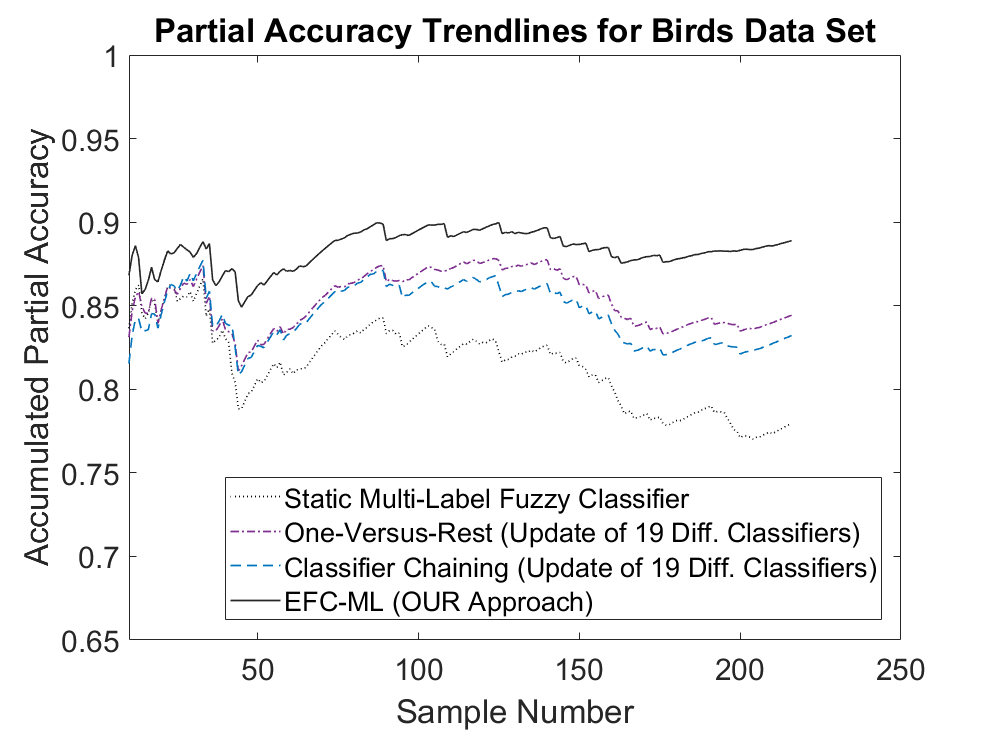}
\includegraphics[width=0.46\textwidth]{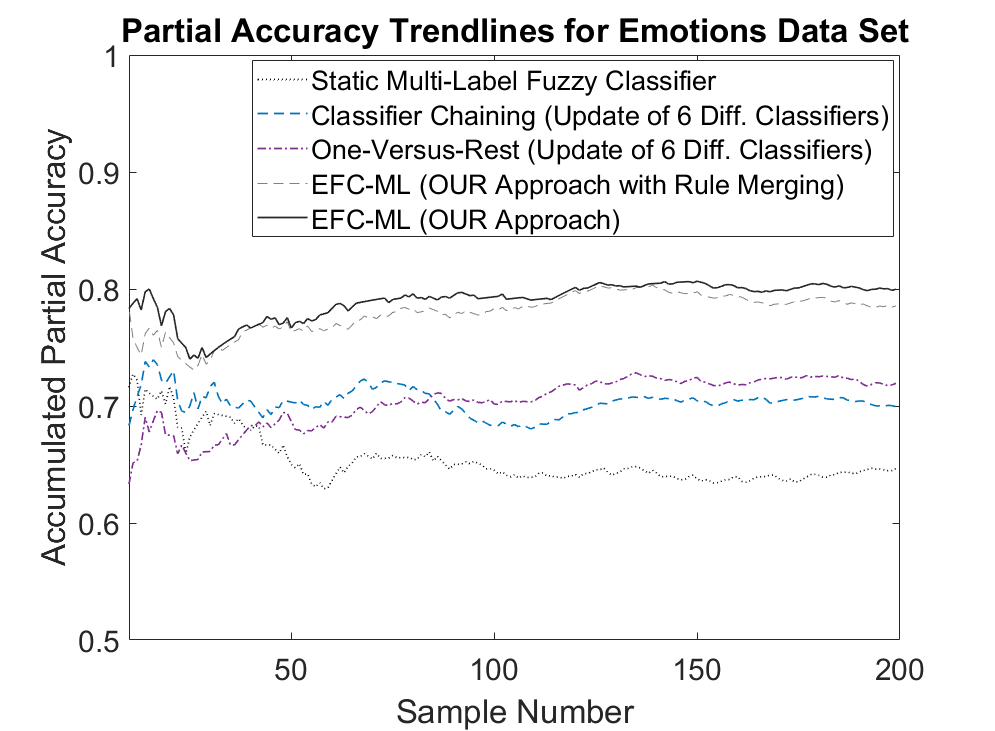}
\includegraphics[width=0.46\textwidth]{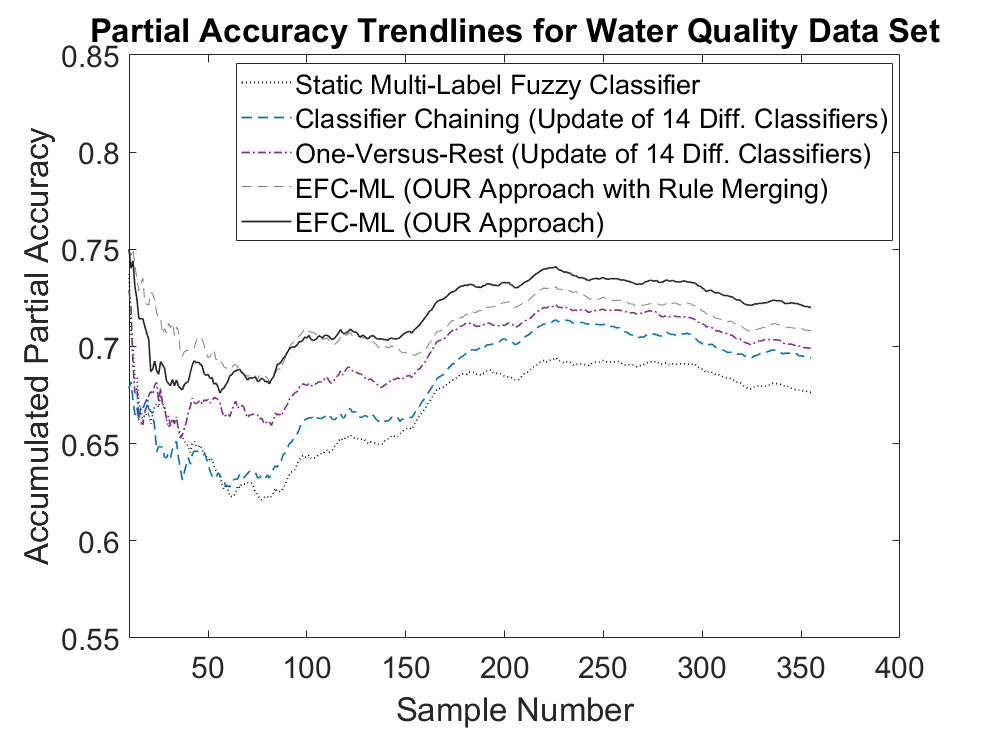}
\includegraphics[width=0.46\textwidth]{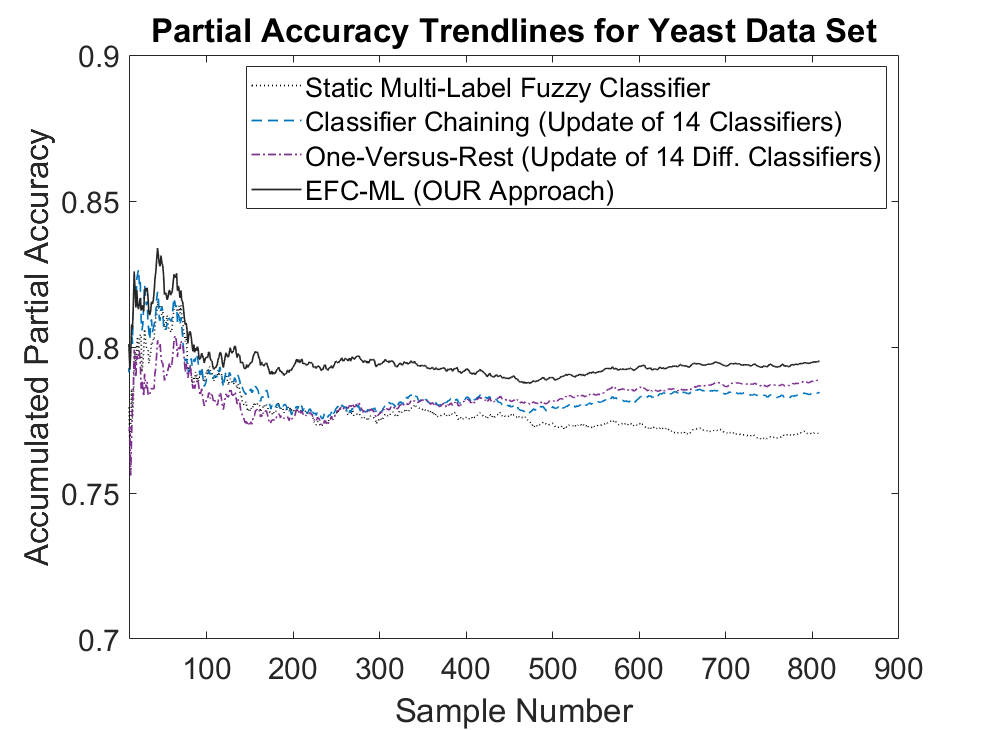}
\caption{Accumulated {\bf partial accuracy} trend lines of the methods under comparison (including our EFC-ML approach, as indicated in the legend) for four data sets as indicated in the titles.}
\label{partial_accuracy_trendlines_smalldatasets}
\end{figure}

Figure \ref{partial_accuracy_trendlines_smalldatasets} compares the partial accuracy trend lines (on the left side) achieved by our evolving multi-label fuzzy classifier on new online samples in comparison with the related SoA works (classifier chaining and one-versus-rest classification): in the upper left plot for the birds data set, in the upper right plot for the emotions data set, in the lower left plot for the water quality data set and in the lower right plot for the yeast data set.
From this figure, it can be seen that our new approach indicated by a solid black line significantly outperforms the related SoA methods for all data streams, showing significantly higher accuracies over more or less the whole streams: in the case of birds around 5\% more accuracy can be gained, in the case of emotions around 10\% more accuracy can be gained anytime. This is a remarkable result, as our approach only requires a single fuzzy partition to explain the granular multi-label structure within the rule antecedents, whereas classifier chaining and one-versus-rest classification implicitly embed $K$ different classifiers with different rule antecedents, with $K$ being the number of different class labels (as indicated in the legends of the figures), which makes them non-transparent and difficult to interpret.
The outperformance is basically due to the embedded incremental correlation-based learning (only possible in the case of a joint antecedent space within a single model), because classifier chaining and one-versus-rest classification variants employ a sole RFWLS estimator for consequent updating.
The static multi-label fuzzy classifier, trained only on the training set portion, but not further updated with any new test samples, achieves the worst performance for all data streams (dotted black markers), which indicates that a refinement and update of the classifiers with new data is necessary, simply because some class label combinations (i.e. multi-label groups) were not included in the initial training data, but occurred dynamically (newly) in the test data set. The likelihood of such an occurrence typically increases with the number of labels, as this increases the number of label combinations exponentially.

\begin{figure}[t]
\centering
\includegraphics[width=0.46\textwidth]{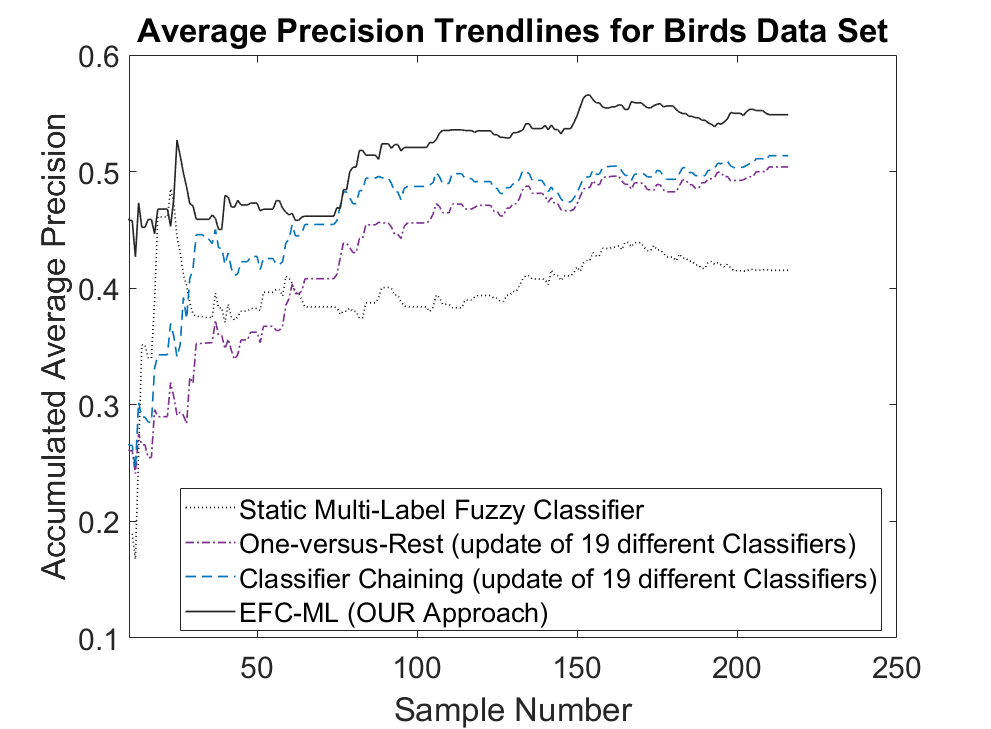}
\includegraphics[width=0.46\textwidth]{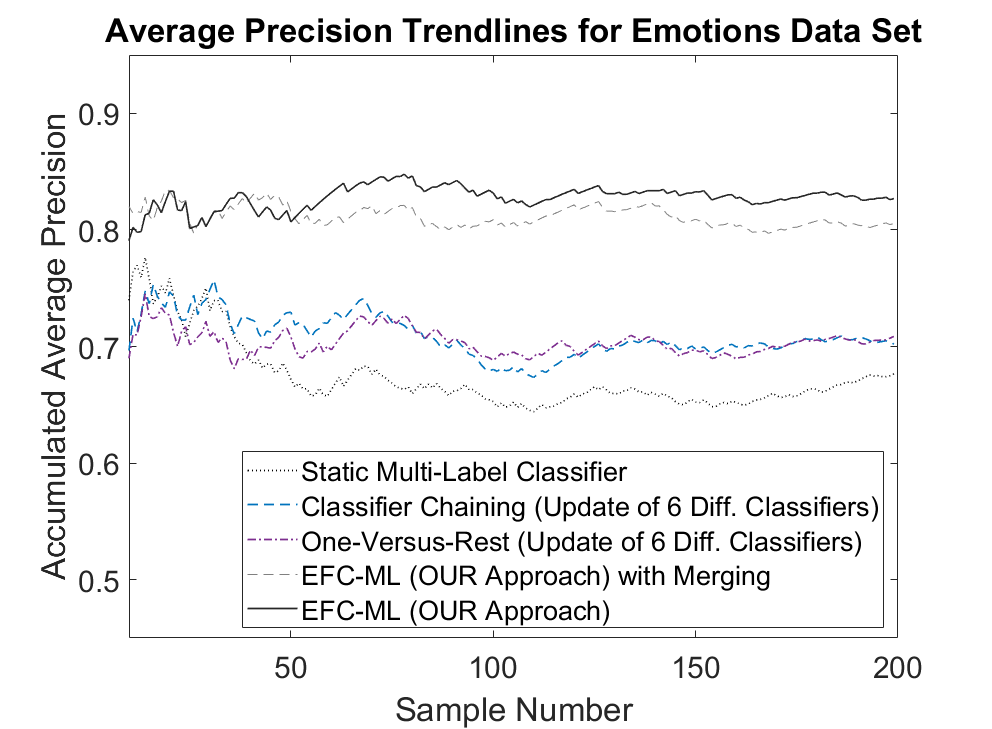}
\includegraphics[width=0.46\textwidth]{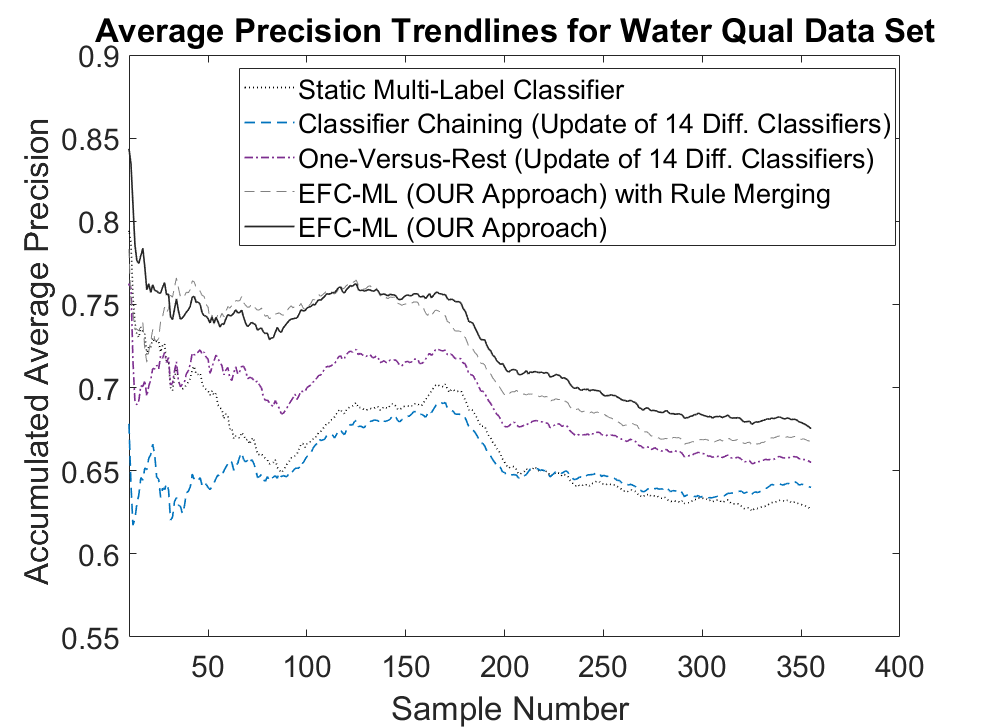}
\includegraphics[width=0.46\textwidth]{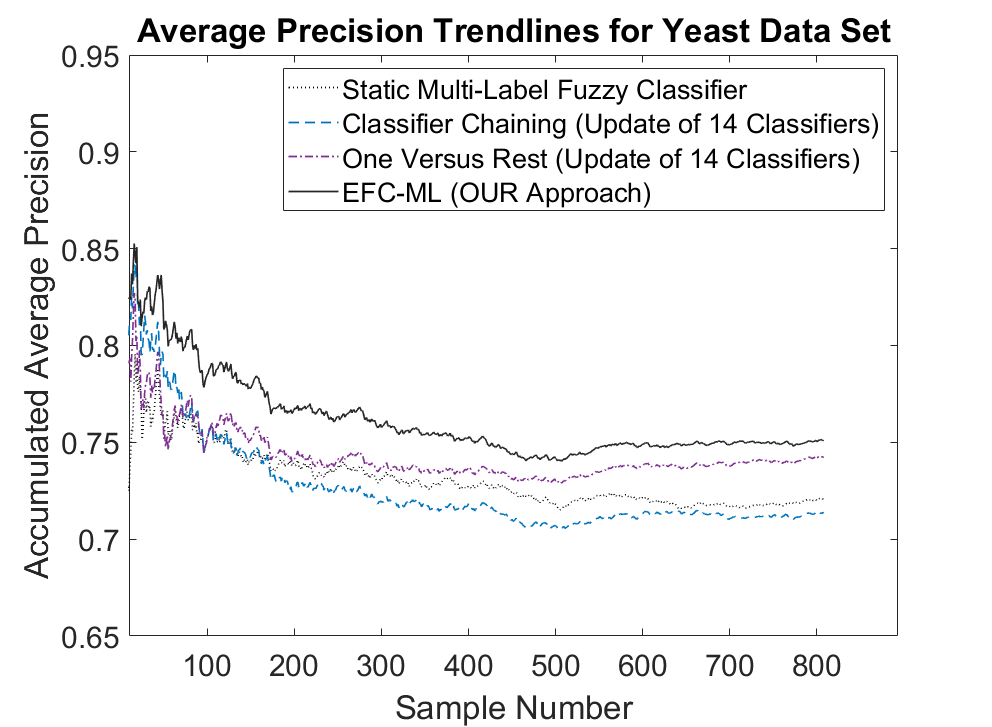}
\caption{Accumulated {\bf average precision} trend lines of the methods under comparison (including our EFC-ML approach, as indicated in the legend) for four data sets as indicated in the titles.}
\label{average_precision_trendlines_smalldatasets}
\end{figure}

A similar picture (out-performance of the related SoA methods and the worst performance of static multi-label fuzzy classifiers) can be observed when inspecting the average precision trend lines as shown in Figure \ref{average_precision_trendlines_smalldatasets}.
The only difference to the partial accuracy trend line tendencies is that classifier chaining is even worse than our static multi-label fuzzy classifiers in the case of the water quality and yeast data sets, despite permanently updating the structures and consequent parameters of the chained classifiers.

\begin{figure}[t]
\centering
\includegraphics[width=0.46\textwidth]{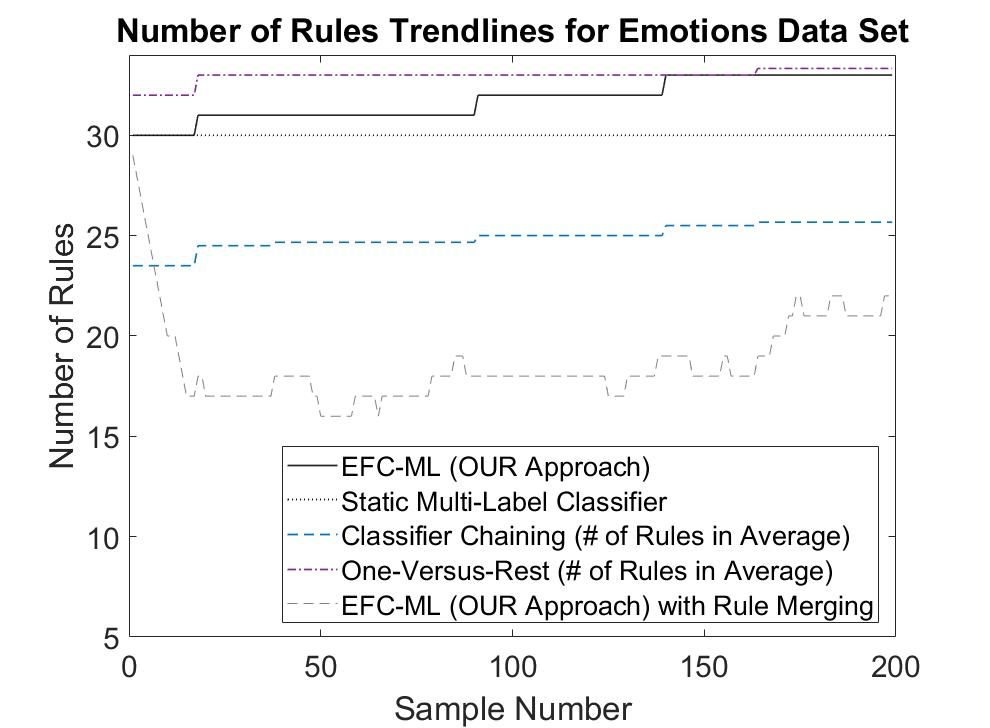}
\includegraphics[width=0.46\textwidth]{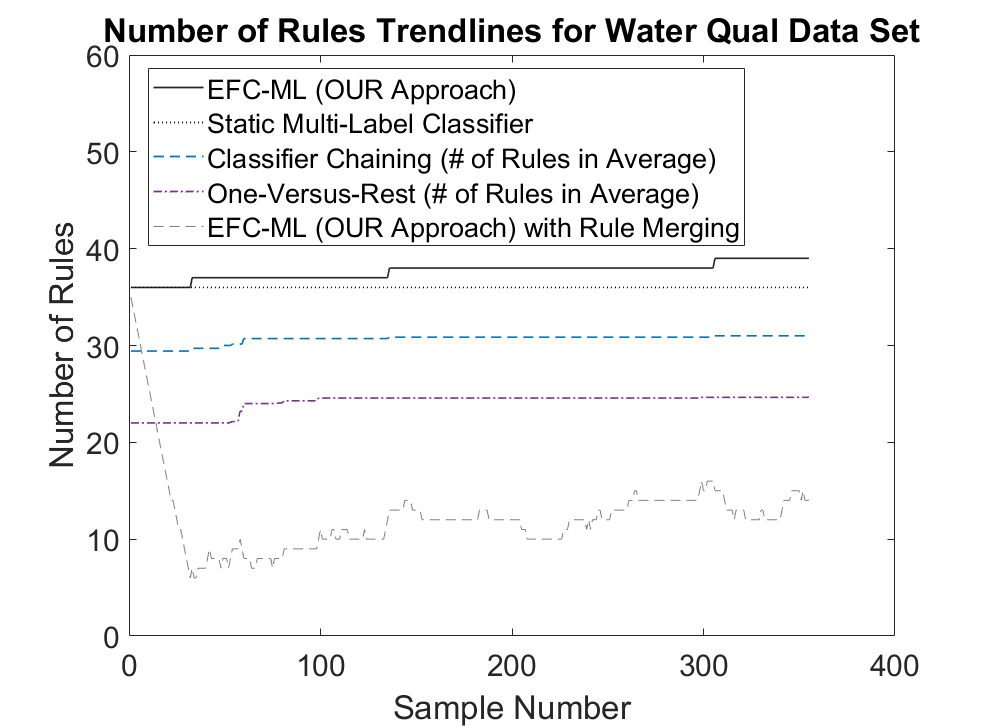}
\caption{Trend lines of the number of rules over stream samples for the emotions (left) and water quality (right) data sets.}
\label{number_rules_trendlines_smalldatasets}
\end{figure}

We also examined the trend in the number of rules (on average for all classifiers for the two related works) for the emotions and water quality data sets. These are shown in Figure \ref{number_rules_trendlines_smalldatasets}.
It can be seen that the embedded rule merging approach in our EFC-ML method can significantly reduce the number of rules (down to around 17 and 10, respectively), while not losing too much partial accuracy and average precision, as shown in Figures \ref{partial_accuracy_trendlines_smalldatasets} and \ref{number_rules_trendlines_smalldatasets}, which may open up some possibilities for rule (base) interpretation and knowledge gaining.
The related classifier variants show a similar or even a lower number of rules, but this number is an average one over all internal classifiers (one per class), which thus has to be multiplied by the number of classes the data sets contain to retrieve the number of rules in total. This yields classifiers containing significantly more than 100 rules, which cannot be realistically read/inspected with a reasonable effort.

\begin{figure}[tp]
\centering
\includegraphics[width=0.46\textwidth]{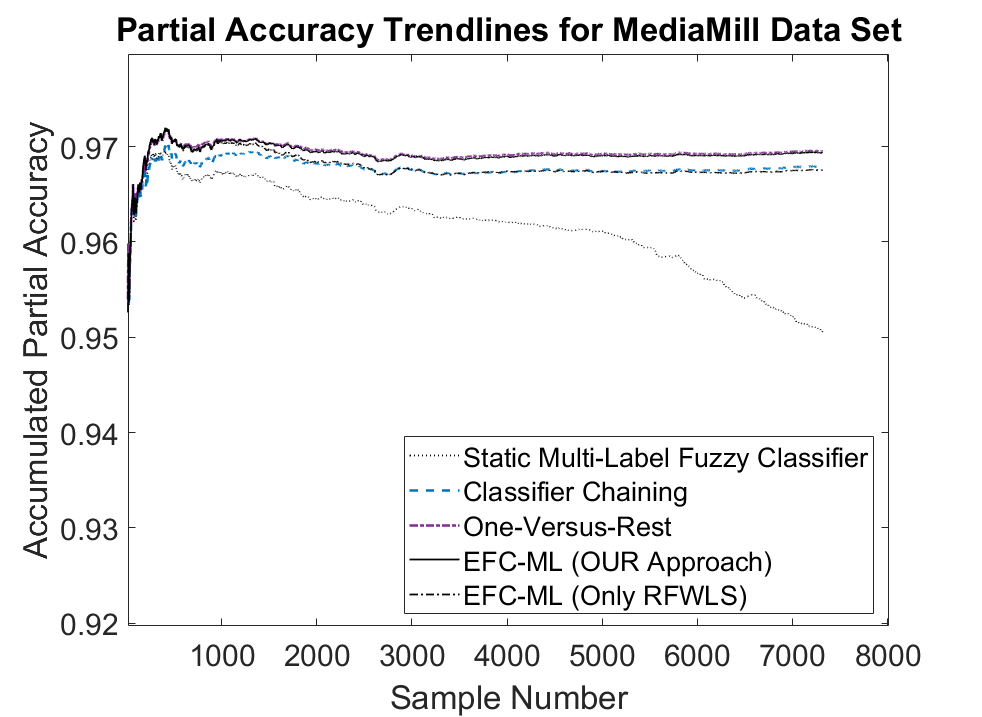}
\includegraphics[width=0.46\textwidth]{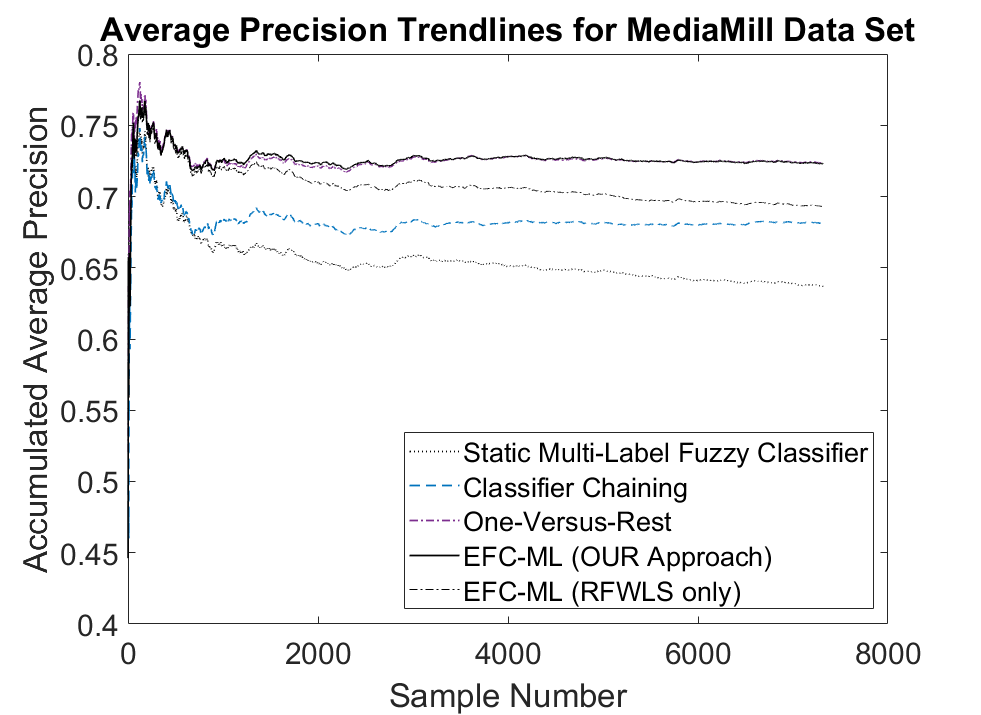}
\caption{Accumulated {\bf partial accuracy} (left) and {\bf average precision} (right) trend-lines of the methods under comparison (including our EFC-ML approach, as indicated in the legend) for the mediamill data set.}
\label{trendlines_mediamill}
\end{figure}

Figure \ref{trendlines_mediamill} shows the partial accuracy (left) and the average precision trend lines achieved on the online mediamill data set embedding a wider scale of labels (101).
The trend lines of the static multi-label fuzzy classifiers show significantly deteriorating curves, especially in the case of the partial accuracy
and this declining trend becomes steeper the longer the stream progresses. This clearly indicates certain dynamics in this data set, which demands a classifier update. Obviously, all of the evolving, incremental classifiers can hold partial accuracies as well as average precision on a much higher level than a static classifier, with our EFC-ML approach and one-versus-rest more or less performing equally (a bit higher average precision could be achieved at the beginning of the stream with our approach) and both clearly outperforming the classifier chaining strategy. Furthermore, we also show the trend line when only performing RFWLS for consequent parameter updates in EFC-ML (and no incremental correlation-based update step), which then leads to a worse performance, more or less ending up the same as classifier chaining. This confirms the usefulness of correlation-based learning to increase robustness.

\subsection{Model Updates with Actively Selected Samples}
Next, we studied the performance trend lines in terms of the partial accuracy of our EFC-ML method in the case when only a limited number of labels is available, as is typically the case when the effort required by operators to provide annotation feedback for each single sample becomes too great.
Therefore, we operated on two levels and in two stages. In the first stage, we only included the novelty and output uncertainty-based criteria for sample selection, as these are convenient concepts in the (online, evolving) active learning community. In the second stage, we also included our new concept regarding parameter uncertainty, which aims to increase sample significance and the robustness of the solution, to see its additional effect on the performance trend lines.
The two levels correspond to labels-based and samples-based active learning in terms of exhausting a pre-defined budget (see Section \ref{evaluation_measures}), where we used 30\%, 20\% and 10\% of a maximally allowed budget the expert has available for labelling the data.

\begin{figure}[t]
\centering
\includegraphics[width=0.46\textwidth]{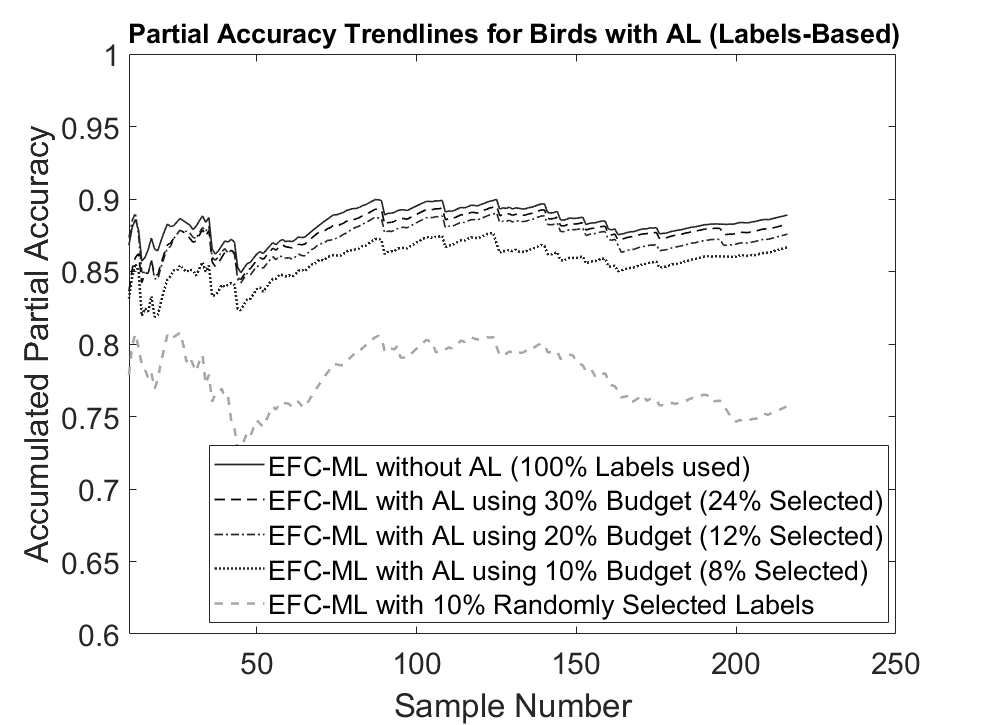}
\includegraphics[width=0.46\textwidth]{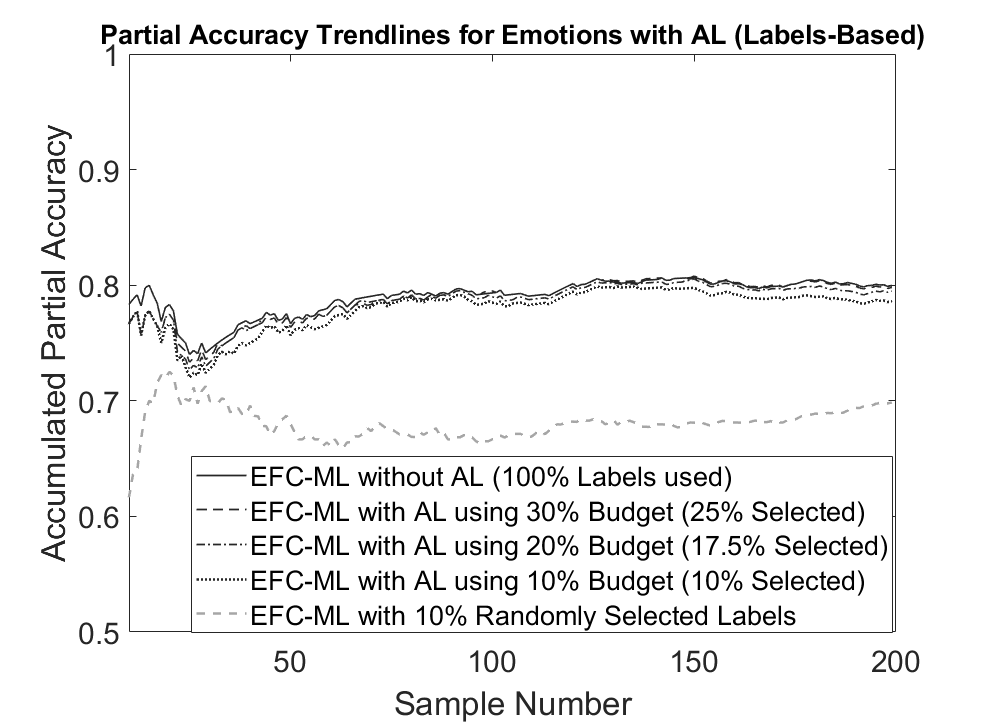}
\includegraphics[width=0.46\textwidth]{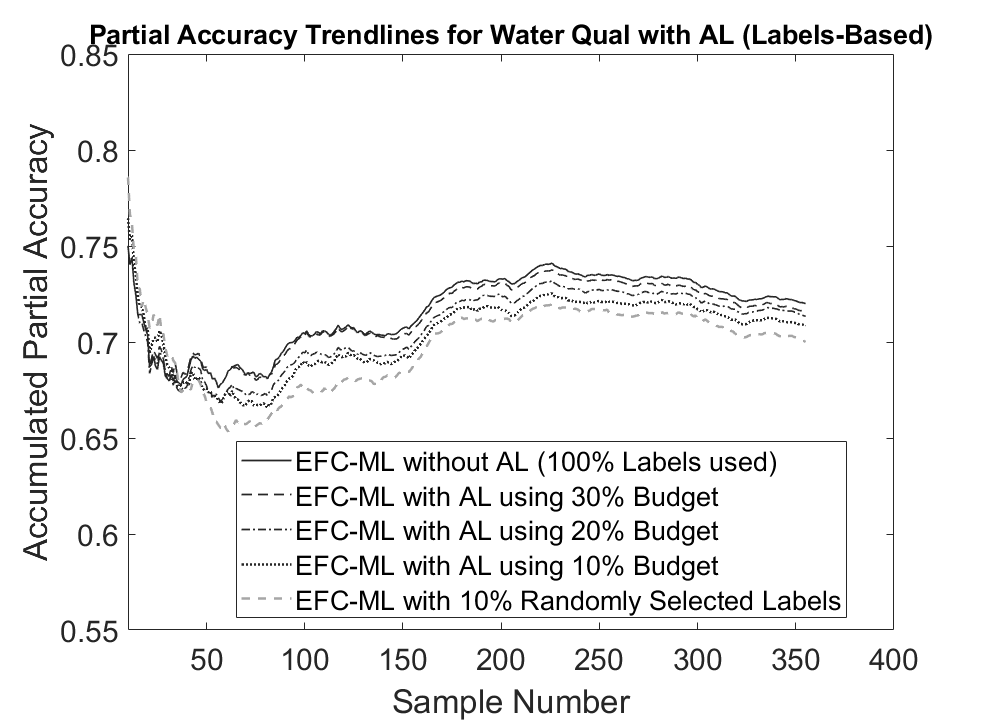}
\includegraphics[width=0.46\textwidth]{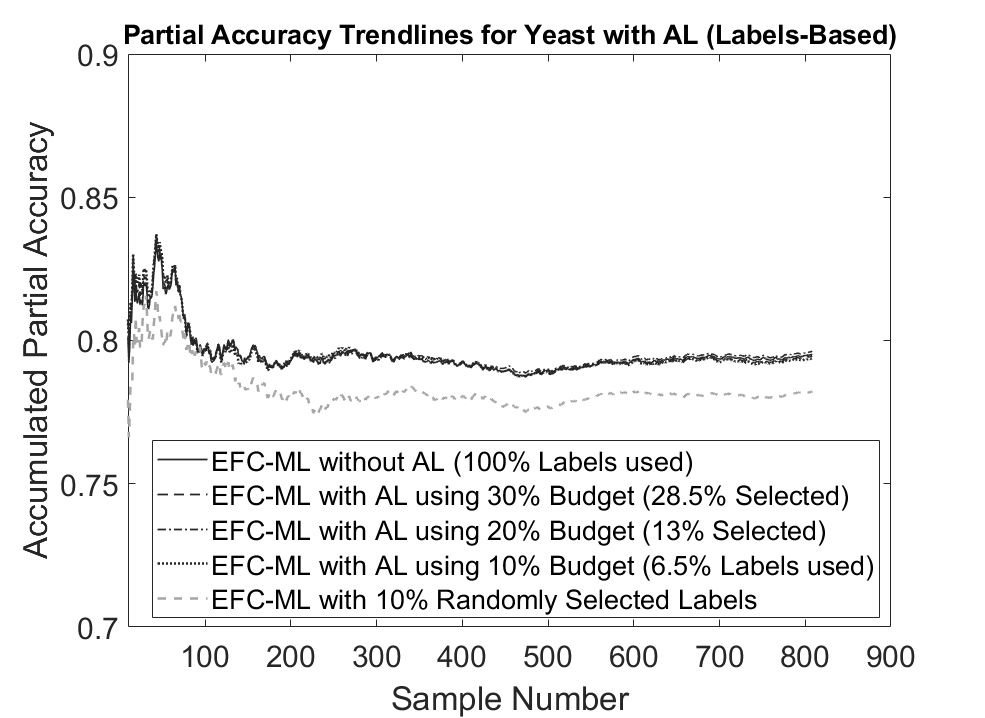}
\caption{Accumulated partial accuracy trend lines for four data sets as indicated in the titles when using online active learning technique in combination with a {\bf labels-based} budget.}
\label{partial_accuracy_trendlines_smalldatasets_oAL_labelsbased}
\end{figure}

Figure \ref{partial_accuracy_trendlines_smalldatasets_oAL_labelsbased} shows the partial accuracy trend lines on four data sets achieved by online sample selection with the labels-based budget in the first stage, which restricts the number of labels selected so far to a maximal percentage subject to the number of labels seen so far (equal to $KN$).
From this figure, it can be seen that for the emotions and yeast data sets, the performance decrease is negligible compared to a full update (solid dark lines) when decreasing the budget to 20\% and even 10\%, whereas an even lower percentage is mostly selected, as indicated in the braces of the legend.
This is remarkable, as 10\% random selection results in significantly lower accuracy trend lines, which are close to those produced by the static multi-label classifier. Hence, the embedded single-pass selection strategy significantly contributes to a better performance.
In the case of the birds and water quality data sets, there is a significant drop in performance when the budget decreases from 20\% to 10\% (but not between 100\% and 20\%), where 8\% and 10\% of the samples are selected. This shows the efficiency of our online active learning method in significantly lowering the annotation effort (reducing it by 80\% or even 90\%) while maintaining a high level of performance close to the full update case.

\begin{figure}[t]
\centering
\includegraphics[width=0.46\textwidth]{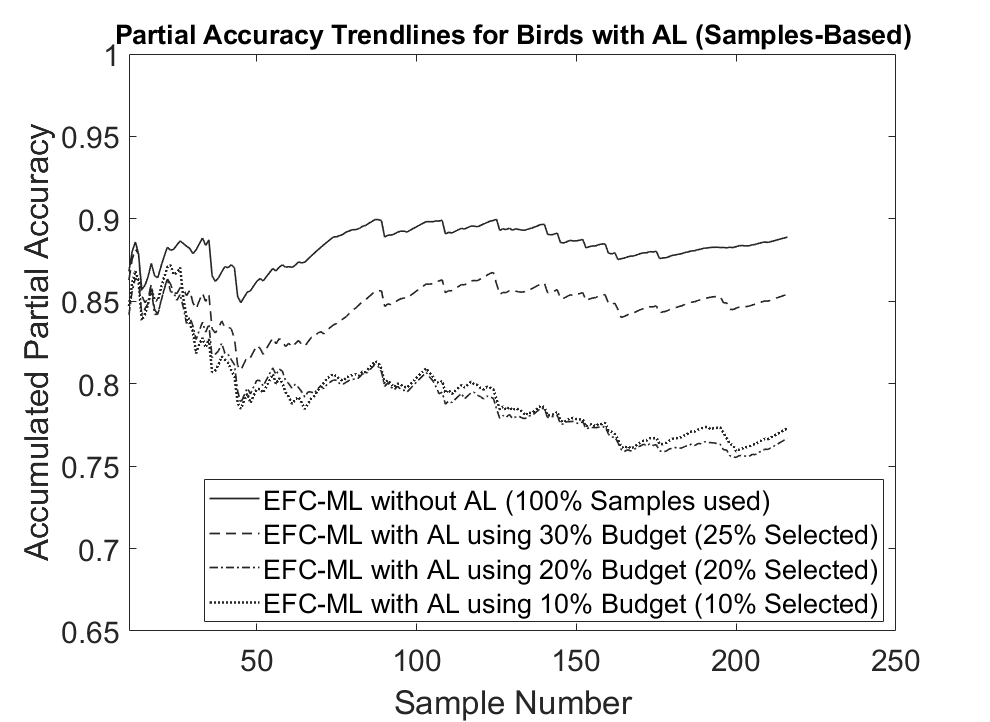}
\includegraphics[width=0.46\textwidth]{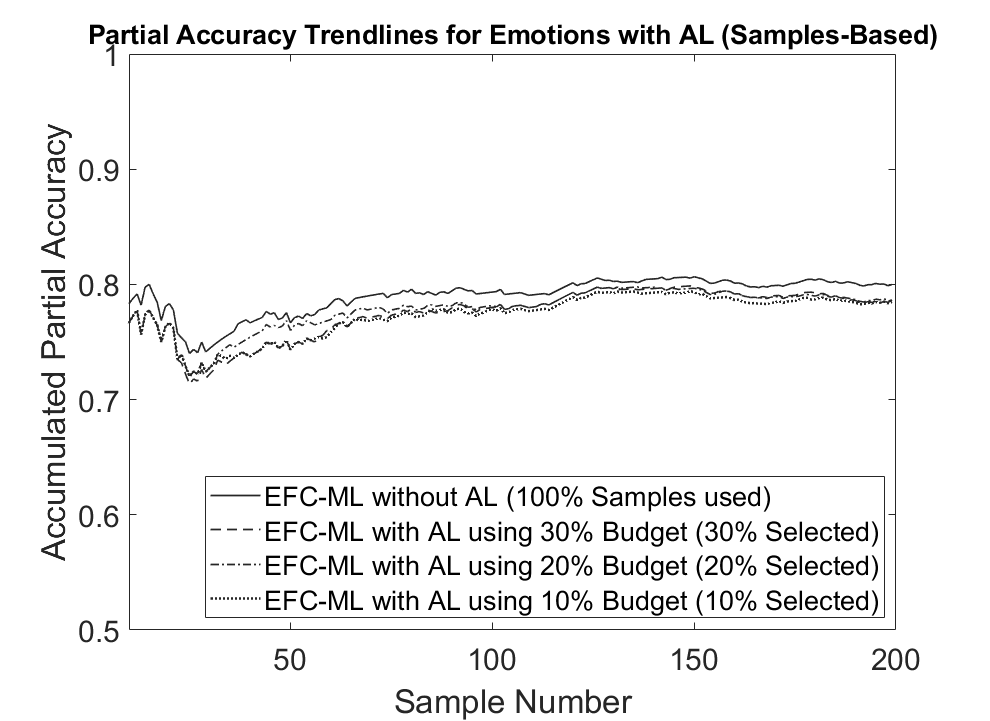}
\includegraphics[width=0.46\textwidth]{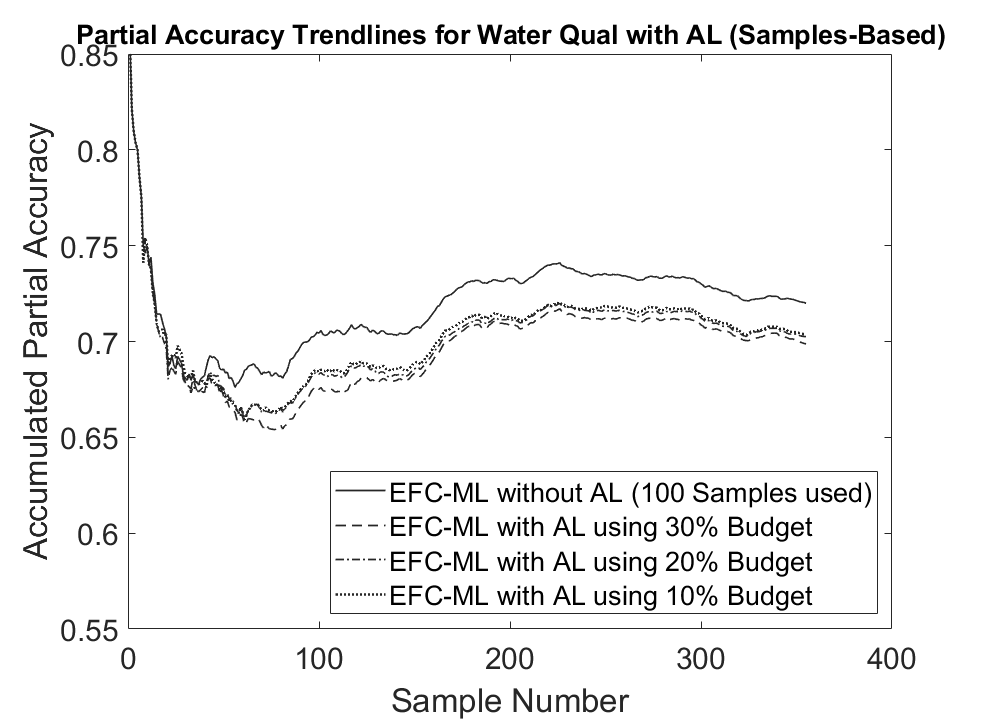}
\includegraphics[width=0.46\textwidth]{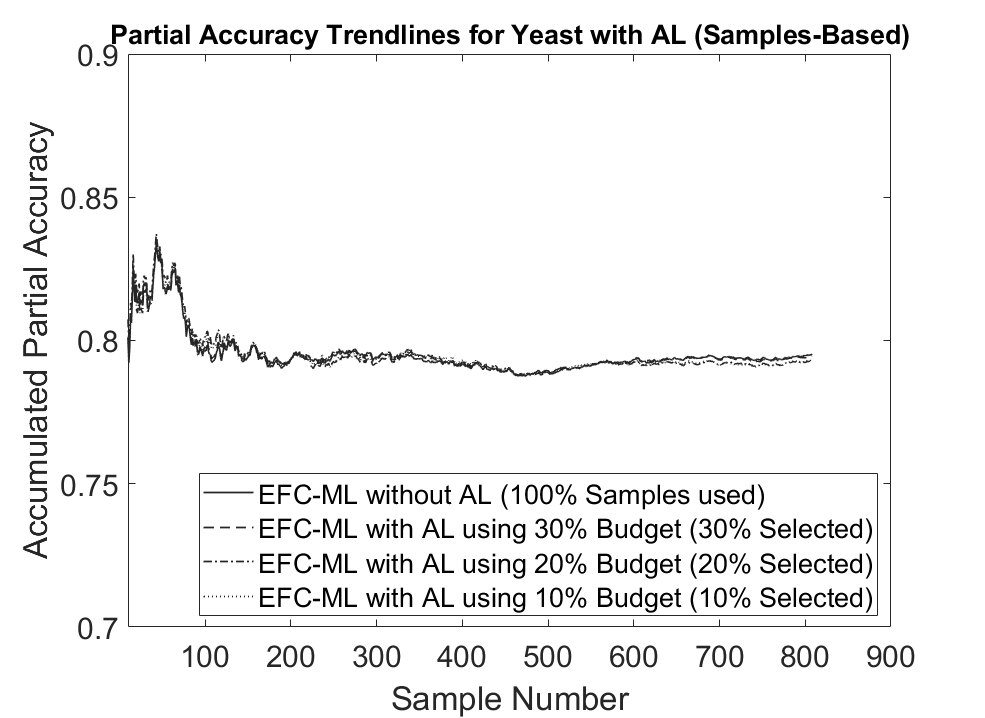}
\caption{Accumulated partial accuracy trend lines for four data sets as indicated in the titles when using online active learning technique in combination with a {\bf samples-based} budget.}
\label{partial_accuracy_trendlines_smalldatasets_oAL_samplesbased}
\end{figure}

On the other hand, the labels-based budget can be seen as an optimistic budget, where it is assumed that a user can provide the partial ground truth to the particular labels requested (0 or 1) without inspecting the other labels of the respective samples selected.
A more pessimistic case from the annotation effort point of view, however, would be that once a sample is selected, the user has to inspect all its labels and provide feedback on them, thus she/he always has to label the complete sample.
Figure \ref{partial_accuracy_trendlines_smalldatasets_oAL_samplesbased} shows the partial accuracy trend lines for the sample-based budget cases.
The effect of the samples-based versus the labels-based budget is negligible on the emotions and yeast data set, as both show nearly the same performance trends as when using all samples (and labels) for updating. For the water quality data set, there is a slight performance drop of around 3\% between a full model update and an update with 30\% of the data, whereas there is no significant further drop when further reducing the budget to 20\% and 10\%, respectively. For the birds data set, obviously 30\% of the data stream is needed for the classifier update to have a higher level of partial accuracy than the static classifier, also compare with Figure \ref{partial_accuracy_trendlines_smalldatasets}.

In the second stage, we added parameter instability/uncertainty as an additional sample selection criterion (as described in Section \ref{instability_paramspace}) to the 10\% budget case to check whether a performance improvement can be obtained on this small, economic budget.
\begin{figure}[t]
\centering
\includegraphics[width=0.46\textwidth]{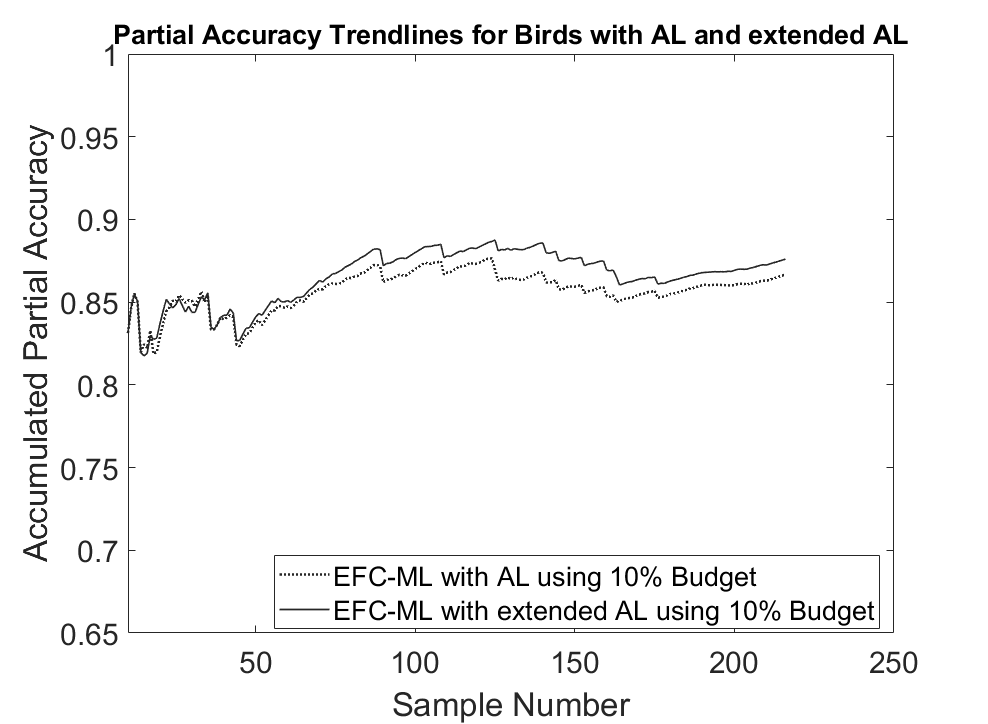}
\includegraphics[width=0.46\textwidth]{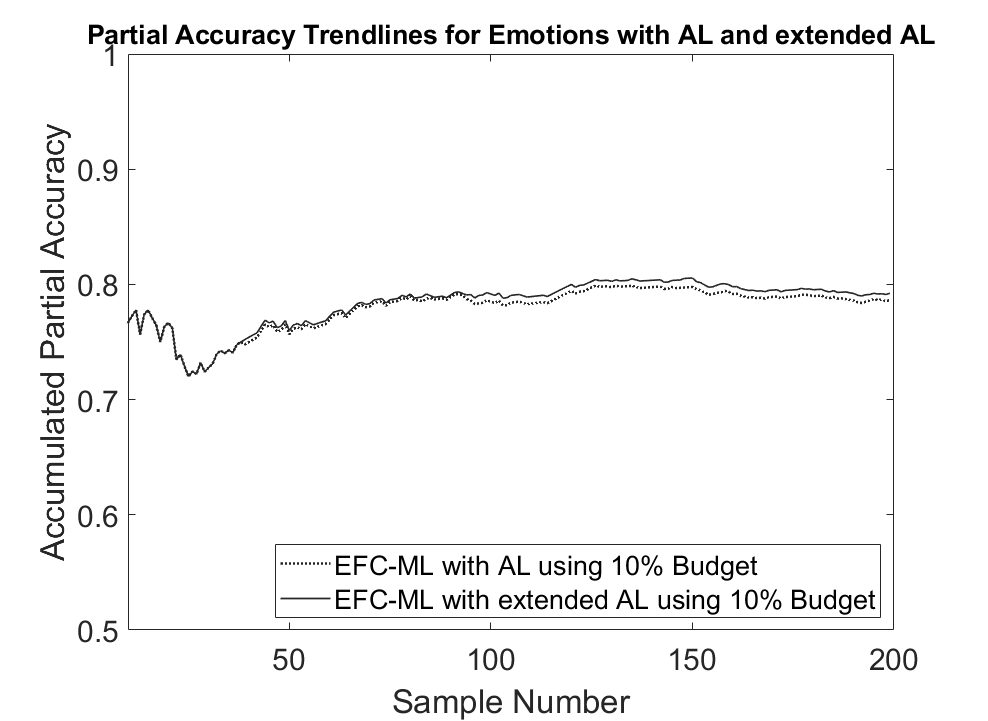}
\includegraphics[width=0.46\textwidth]{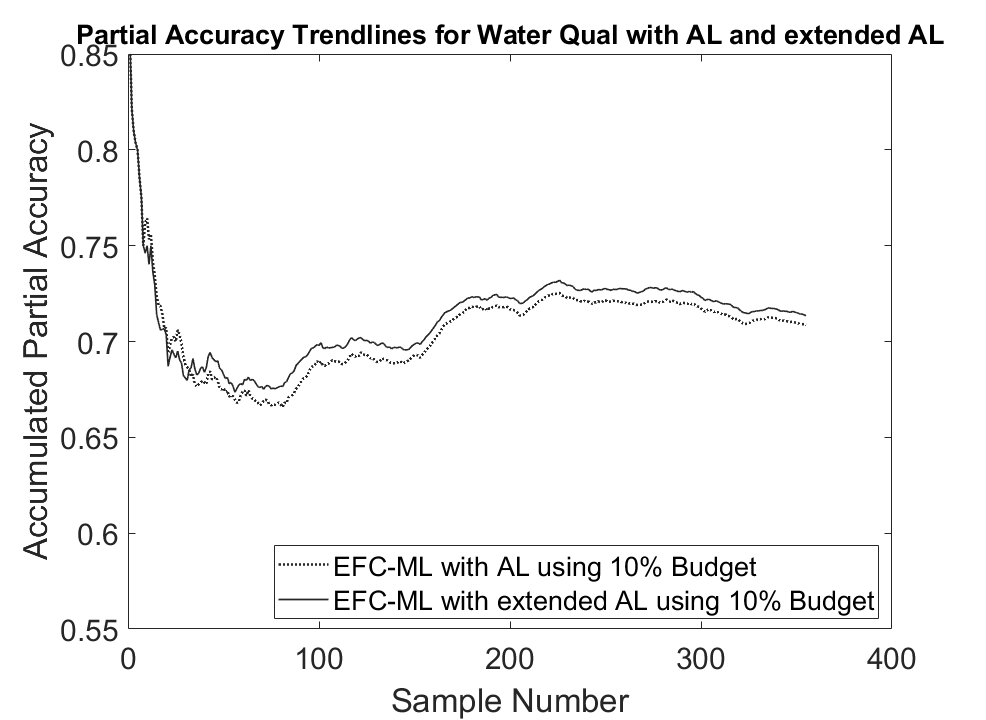}
\includegraphics[width=0.46\textwidth]{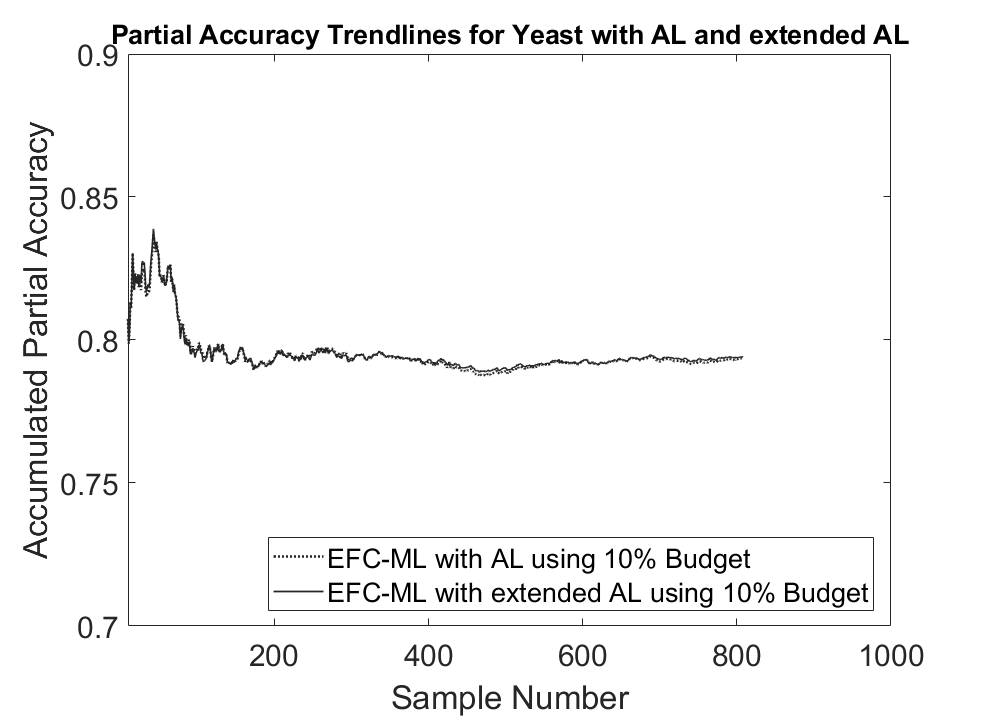}
\caption{Accumulated partial accuracy trend lines for four data sets as indicated in the titles when using extended online active learning technique employing parameter uncertainty criterion.}
\label{partial_accuracy_trendlines_smalldatasets_oAL_labelsbased_ext}
\end{figure}
Figure \ref{partial_accuracy_trendlines_smalldatasets_oAL_labelsbased_ext} shows the performance improvement in the partial accuracy trend lines for the birds, emotions and water quality data sets, with the highest increase being for the birds data set, which has by far the highest input dimensionality of all the data sets, which in turn leads to the highest number of consequent parameters in the rules to be learned from the data. Hence, increasing parameter stability seems to be more important for a larger set of parameters, which is somehow intuitively expected due to the well-known curse of dimensionality effect.

\begin{figure}[tp]
\centering
\includegraphics[width=0.46\textwidth]{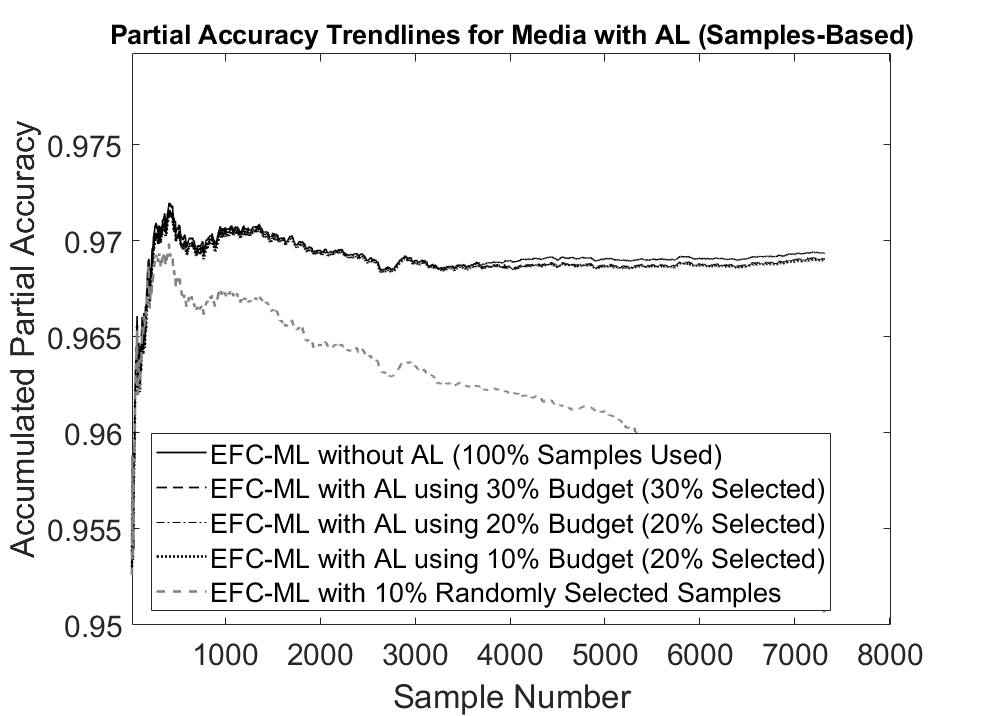}
\includegraphics[width=0.46\textwidth]{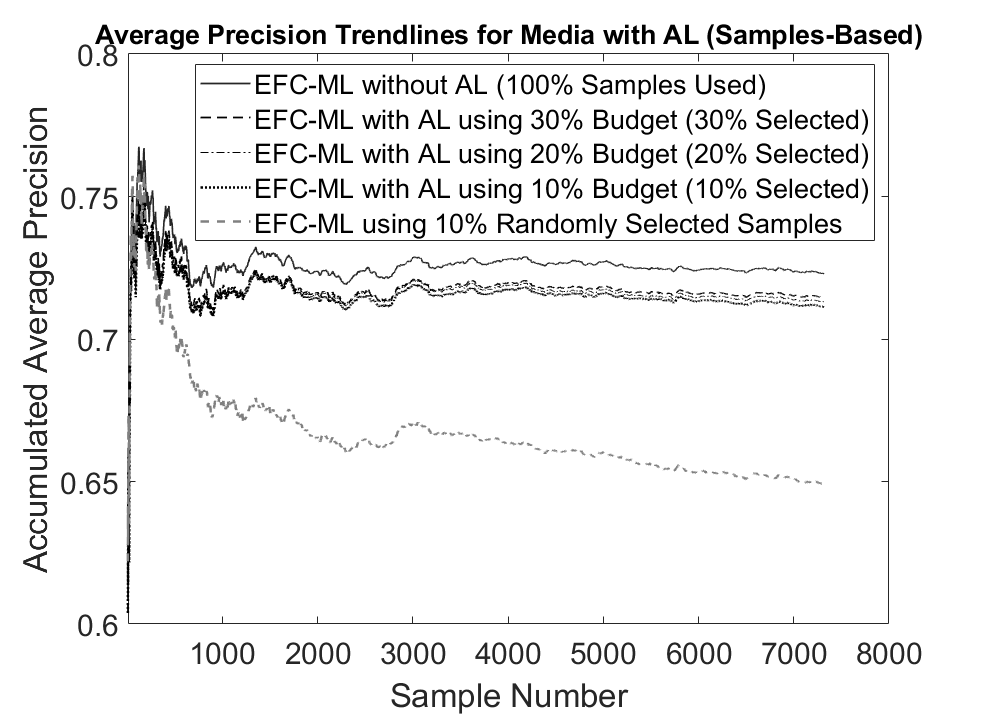}
\caption{Accumulated {\bf partial accuracy} (left) and {\bf average precision} (right) trend lines of the methods under comparison for the mediamill data set when using online active learning technique in combination with a {\bf samples-based} budget.}
\label{trendlines_OAL_mediamill}
\end{figure}

Figure \ref{trendlines_OAL_mediamill} shows the results of the online active sample selection strategy with the samples-based budget applied on the mediamill data set and indicates similar behavior as for the other data sets: partial accuracy does not significantly drop at all, not even when only 10\% of the samples are used for classifier updates, whereas the average precision curve drops slightly between a budget of 100\% and 30\%.
It is remarkable that, when selecting 10\% of the samples randomly, almost the same curve trends as for the static multi-fuzzy classifiers are achieved. This indicates that an active sample selection is necessary to keep the classifier's performance on a high level while trying to be as economic as possible.

Finally, we performed an empirical comparison on the computational efficiency of the methods, as this is a key issue when evolving the classifier from data streams, which are coming in and thus need to be processed on-line with a certain speed and usually based on single samples. Therefore, in Table \ref{computational_efficiency_comparison} we show the update speed of our method in comparison with the related SoA works (one-versus-rest and classifier chaining) for a single sample in average on the same five data sets we have used for accuracy comparisons above.
\begin{table}[t]
\begin{center}
\caption{{\small Comparison of computational efficiency of the methods in terms of update speed on single samples in average (measured in seconds).}}
\label{computational_efficiency_comparison}
{\small
\begin{tabular}{|l|c|c|c|}
\hline
Data Set & EFC-ML (our) & One-Versus-Rest & Classifier Chaining \\
\hline \hline
Birds & 0.289 & 0.868 & 1.078 \\
\hline
Emotions & 0.070 & 0.107  & 0.172  \\
\hline
Water Quality & 0.033 & 0.093 & 0.324 \\
\hline
Yeast & 0.063 & 0.147 & 0.171 \\
\hline
Mediamill & 0.41 & 0.94 & 1.56 \\
\hline
\end{tabular} }
\end{center}
\end{table}
From this table, it can be recognized that our method performs at least three times faster than the classifier chaining method and at least two times faster than the one-versus-rest approach for all data sets (except for the emotions data set), thus being able to handle a three times or two times higher frequency processing speed than these related approaches. This is remarkable, as we have used here each data sample for model update. However, according to the empirical study above using on-line active learning (oAL) methodology, a significant reduction of samples required for the update can be achieved down to around 10\% with only little loss on prediction accuracy. This finally means that we achieve a 20/30 times faster processing speed than the related methods (which are not equipped with oAL).
In absolute numbers, our EFC-ML classifier can cope with 1 Hz frequency in the case of birds and mediamills data sets and with 10Hz frequency processing in the case of the other three data sets, when using all data samples for model update, and 10 Hz and 100 Hz, respectively, when using the oAL methodology.

\section{Conclusion}
\label{conclusion}
We proposed a novel evolving fuzzy classifier for multi-label stream classification tasks (EFC-ML), which is able to self-adapt its rule structure on-the-fly and to take into account possible correlations among the class labels due to an advanced incremental optimization of consequent parameters with guaranteed convergence due to the convexity of the local optimization problems (one per rule). Locality is achieved through a weighted formulation of the optimization problems and is important to avoid class masking problems and to achieve smaller optimization problems. The antecedent learning takes place in the product space and generates one rule partition for all labels together, which allows a compact and transparent knowledge view about feature-class dependencies and thus enables interpretable insights. An online active learning strategy is integrated to significantly reduce the annotation effort for users when providing class labels. It is based on a combination of novelty content-based selection (responsible for knowledge expansion), classifier output uncertainty (responsible for sharpening the decision boundaries) and classifier parameter uncertainty-based selection (responsible for increasing robustness), and integrates a maximal upper allowed budget for labeling which is not to be exceeded.
The results showed a significantly improved performance compared to related multi-label classification strategies such as one-versus-rest classification and classifier chaining, based on five multi-label classification data sets from the MULAN repository. Online active learning could significantly reduce the required class labels to around 10\%, while still leading to similar accuracy trend lines as when using fully labelled samples.

Possible future work already in plan concerns the integration of (human) label uncertainty and external expert knowledge (in form of rules for certain class constellations) into the evolving update algorithms of the multi-label classifiers in order to achieve a hybrid learning approach with cognitive aspects, as well as the improvement of the transparency of the rules in the classifiers by performing a rule length reduction through incremental dimension reduction. The latter is also for the purpose to reduce the curse of dimensionality effect, which is often present in multi-labelled data sets containing a large number of input features. This should thus further increase the predictive performance leading to higher accuracy trend lines, while making the rule-base more compact.

\section*{Acknowledgements}
This work has been supported by the Austrian Science Fund (FWF), contract number
P32272-N38 (acronym IL-EFS), and by the COMET-K2 Center of the Linz Center of Mechatronics (LCM) funded by the Austrian federal government and the federal state of Upper Austria.

\section*{References}


\begin{thebibliography}{10}

\bibitem{AngelovFilevKasabov10}
P.~Angelov, D.~Filev, and N.~Kasabov.
\newblock {\em Evolving Intelligent Systems --- Methodology and Applications}.
\newblock John Wiley \& Sons, New York, 2010.

\bibitem{AngelovGu18}
P.~Angelov and X.~Gu.
\newblock Deep rule-based classifier with human-level performance and
  characteristics.
\newblock {\em Information Sciences}, 463--464:196--213, 2018.

\bibitem{AngelovGuPrincipe18}
P.~Angelov, X.~Gu, and J.C. Principe.
\newblock Autonomous learning multimodel systems from data streams.
\newblock {\em IEEE Transactions on Fuzzy Systems}, 26(4):2213--2224, 2018.

\bibitem{AngelovGuPrincipe17}
P.~Angelov, X.~Gu, and J.C. Principe.
\newblock A generalized methodology for data analysis.
\newblock {\em IEEE Transactions on Cybernetics}, 48(10):2981--2993, 2018.

\bibitem{AngelovLughoferZhou09}
P.~Angelov, E.~Lughofer, and X.~Zhou.
\newblock Evolving fuzzy classifiers using different model architectures.
\newblock {\em Fuzzy Sets and Systems}, 159(23):3160--3182, 2008.

\bibitem{AngelovYager12}
P.~Angelov and R.~Yager.
\newblock A new type of simplified fuzzy rule-based system.
\newblock {\em International Journal of General Systems}, 41(2):163--185, 2012.

\bibitem{AngelovZhou08}
P.~Angelov and X.~Zhou.
\newblock Evolving fuzzy-rule-based classifiers from data streams.
\newblock {\em IEEE Transactions on Fuzzy Systems}, 16(6):1462--1475, 2008.

\bibitem{Angelov12}
P.P. Angelov.
\newblock {\em Autonomous Learning Systems: From Data Streams to Knowledge in
  Real-time}.
\newblock John Wiley \& Sons, New York, 2012.

\bibitem{AngelovFilev04}
P.P. Angelov and D.~Filev.
\newblock An approach to online identification of {T}akagi-{S}ugeno fuzzy
  models.
\newblock {\em IEEE Transactions on Systems, Man and Cybernetics, part B:
  Cybernetics}, 34(1):484--498, 2004.

\bibitem{BehniafarNowrooziShahriari18}
M.~Behniafar, A.~Nowroozi, and H.R. Shahriari.
\newblock A survey of anomaly detection approaches in internet of things.
\newblock {\em The {ISC} International Journal of Information Security},
  10(2):79--92, 2018.

\bibitem{SouzaLughoferNC20}
P.V. de~Campos~Souza and E.~Lughofer.
\newblock An evolving neuro-fuzzy system based on uni-nullneurons with advanced
  interpretability capabilities.
\newblock {\em Neurocomputing}, 451(231--251), 2021.

\bibitem{DonmezCarbonell10}
P.~Donmez and J.G. Carbonell.
\newblock From active to proactive learning methods.
\newblock In J.~Koronacki, Z.W. Ras, S.T. Wierzchon, and J.~Kacprzyk, editors,
  {\em Advances in Machine Learning I}, volume 262 of {\em Studies in
  Computational Intelligence}, pages 97--120. Springer, Berlin Heidelberg,
  2010.

\bibitem{EfronHastieJohnstoneTibshirani04}
B.~Efron, T.~Hastie, I.~Johnstone, and R.~Tibshirani.
\newblock Least angle regression.
\newblock {\em The Annals of Statistics}, 32(2):407--451, 2004.

\bibitem{Frieden04}
B.R. Frieden.
\newblock {\em Science from Fisher Information: A Unification}.
\newblock Cambridge University Press, Cambridge, UK, 2004.

\bibitem{GarciaLeiteSkrjanc20}
C.~Garcia, D.~Leite, and I.~Skrjanc.
\newblock Incremental missing-data imputation for evolving fuzzy granular
  prediction.
\newblock {\em IEEE Transactions on Fuzzy Systems}, 28(10):2348--2361, 2020.

\bibitem{GolubVanLoan96}
G.H. Golub and C.F.~Van Loan.
\newblock {\em Matrix Computations (3rd Edition)}.
\newblock John Hopkins University Press, Baltimore, Maryland, 1996.

\bibitem{HerreraCharte18}
F.~Herrera, F.~Charte, A.J. Rivera, and M.J. del Jesus.
\newblock {\em Multilabel Classification: Problem Analysis, Metrics and
  Techniques}.
\newblock Springer, Switzerland, 2018.

\bibitem{HisadaOzawaZhangKasabov10}
M.~Hisada, S.~Ozawa, K.~Zhang, and N.~Kasabov.
\newblock Incremental linear discriminant analysis for evolving feature spaces
  in multitask pattern recognition problems.
\newblock {\em Evolving Systems}, 1(1):17--27, 2010.

\bibitem{JiTangYuYe08}
S.~Ji, L.~Tang, S.~Yu, and J.~Ye.
\newblock Extracting shared subspace for multi-label classification.
\newblock In {\em Proceedings of the 14th {ACM SIGKDD} International Conference
  on Knowledge Discovery and Data Mining}, pages 381--389, 2008.

\bibitem{JohnHagrasCastillo19}
R.~John, H.~Hagras, and O.~Castillo.
\newblock {\em Type-2 Fuzzy Logic and Systems}.
\newblock Springer Verlag, Cham, Switzerland, 2019.

\bibitem{KanginAngelovIglesias15}
D.~Kangin, P.~Angelov, and J.A. Iglesias.
\newblock Autonomously evolving classifier {TEDAC}lass.
\newblock {\em Information Sciences}, 366:1--11, 2016.

\bibitem{KremplKottkeLemaire15}
G.~Krempl, D.~Kottke, and V.~Lemaire.
\newblock Optimised probabilistic active learning ({OPAL}).
\newblock {\em Machine Learning}, 100(2--3):449--476, 2015.

\bibitem{LeiteAndonovskiSkrjancGomide19}
D.~Leite, R.M. Palhares, C.~S. Campos, and Fernando Gomide.
\newblock Optimal rule-based granular systems from data streams.
\newblock {\em IEEE Transactions on Fuzzy Systems}, 28(3):583--596, 2020.

\bibitem{LiWangPavluAslam16}
C.~Li, B.~Wang, V.~Pavlu, and J.~Aslam.
\newblock Conditional bernoulli mixtures for multi-label classification.
\newblock In {\em Proceedings of the International Conference on Machine
  Learning 2016}, pages 2482--2491, 2016.

\bibitem{LughoferPositionPaper13}
E.~Lughofer.
\newblock On-line assurance of interpretability criteria in evolving fuzzy
  systems --- achievements, new concepts and open issues.
\newblock {\em Information Sciences}, 251:22--46, 2013.

\bibitem{LughoferChapter14}
E.~Lughofer.
\newblock Evolving fuzzy systems --- fundamentals, reliability,
  interpretability and useability.
\newblock In P.~Angelov, editor, {\em Handbook of Computational Intelligence},
  pages 67--135. World Scientific, New York, 2016.

\bibitem{Lughofer17}
E.~Lughofer.
\newblock On-line active learning: A new paradigm to improve practical
  useability of data stream modeling methods.
\newblock {\em Information Sciences}, 415--416:356--376, 2017.

\bibitem{LughoferBuchtala13}
E.~Lughofer and O.~Buchtala.
\newblock Reliable all-pairs evolving fuzzy classifiers.
\newblock {\em IEEE Transactions on Fuzzy Systems}, 21(4):625--641, 2013.

\bibitem{LughoferCernudaKindermannPratama14}
E.~Lughofer, C.~Cernuda, S.~Kindermann, and M.~Pratama.
\newblock Generalized smart evolving fuzzy systems.
\newblock {\em Evolving Systems}, 6(4):269--292, 2015.

\bibitem{LughoferKindermann09jour}
E.~Lughofer and S.~Kindermann.
\newblock Sparse{FIS}: Data-driven learning of fuzzy systems with sparsity
  constraints.
\newblock {\em IEEE Transactions on Fuzzy Systems}, 18(2):396--411, 2010.

\bibitem{LughoferRichterNeisslHeidlEitzingerRadauer17}
E.~Lughofer, R.~Richter, U.~Neissl, W.~Heidl, C.~Eitzinger, and T.~Radauer.
\newblock Explaining classifier decisions linguistically for stimulating and
  improving operators labeling behavior.
\newblock {\em Information Sciences}, 420:16--36, 2017.

\bibitem{LughoferMouchaweh19}
E.~Lughofer and M.~Sayed-Mouchaweh.
\newblock {\em Predictive Maintenance in Dynamic Systems --- Advanced Methods,
  Decision Support Tools and Real-World Applications}.
\newblock Springer, New York, 2019.

\bibitem{PedryczSkowronKreinovich08}
W.~Pedrycz, A.~Skowron, and V.~Kreinovich.
\newblock {\em Handbook of Granular Computing}.
\newblock John Wiley \& Sons, Chichester, West Sussex, England, 2008.

\bibitem{PratamaAnavattiLu15}
M.~Pratama, S.~Anavatti, and J.~Lu.
\newblock Recurrent classifier based on an incremental meta-cognitive
  scaffolding algorithm.
\newblock {\em IEEE Transactions on Fuzzy Systems}, 23(6):2048--2066, 2015.

\bibitem{PratamaAnavattiLughoferjour14}
M.~Pratama, S.G. Anavatti, M.J. Er, and E.~Lughofer.
\newblock p{C}lass: An effective classifier for streaming examples.
\newblock {\em IEEE Transactions on Fuzzy Systems}, 23(2):369--386, 2015.

\bibitem{PratamaLuLughoferZhangAnavatti16}
M.~Pratama, J.~Lu, E.~Lughofer, G.~Zhang, and S.~Anavatti.
\newblock Scaffolding type-2 classifier for incremental learning under concept
  drifts.
\newblock {\em Neurocomputing}, 191:304--329, 2016.

\bibitem{PratamaLuZhang16}
M.~Pratama, J.~Lu, and G.~Zhang.
\newblock Evolving type-2 fuzzy classifier.
\newblock {\em IEEE Transactions on Fuzzy Systems}, 24(3):574--589, 2016.

\bibitem{ReadPfahringerHolmes11}
J.~Read, B.~Pfahringer, G.~Holmes, and E.~Frank.
\newblock Classifier chains for multi-label classification.
\newblock {\em Machine Learning Journal}, 85(3), 2011.

\bibitem{SamekMontavon19}
W.~Samek and G.~Montavon.
\newblock {\em Explainable AI: Interpreting, Explaining and Visualizing Deep
  Learning}.
\newblock Springer Nature, Switzerland, 2019.

\bibitem{SkrjancIglesiasLughoferGomide19}
I.~Skrjanc, J.~Iglesias, A.~Sanchis, E.~Lughofer, and F.~Gomide.
\newblock Evolving fuzzy and neuro-fuzzy approaches in clustering, regression,
  identification, and classification: A survey.
\newblock {\em Information Sciences}, 490:344--368, 2019.

\bibitem{SubramanianSureshSundararajan13}
K.~Subramanian, S.~Suresh, and N.~Sundararajan.
\newblock A metacognitive neuro-fuzzy inference system (mcfis) for sequential
  classification problems.
\newblock {\em IEEE Transactions on Fuzzy Systems}, 21(6):1080--1095, 2013.

\bibitem{TungQuekGuan13}
S.W. Tung, C.~Quek, and C.~Guan.
\newblock e{T}2{FIS}: An evolving type-2 neural fuzzy inference system.
\newblock {\em Information Sciences}, 220:124--148, 2013.

\bibitem{WangJiJin13}
L.~Wang, H.-B. Ji, and Y.~Jin.
\newblock Fuzzy passive--aggressive classification: A robust and efficient
  algorithm for online classification problems.
\newblock {\em Information Sciences}, 220:46--63, 2013.

\bibitem{WuShengZhangLi20}
Jian Wu, Victor~S. Sheng, Jing Zhang, Hua Li, Tetiana Dadakova, Christine~Leon
  Swisher, Zhiming Cui, and Pengpeng Zhao.
\newblock Multi-label active learning algorithms for image classification:
  Overview and future promise.
\newblock {\em ACM Comput. Surv.}, 53(2), March 2020.

\bibitem{ZainLughoferPratama17}
C.~Zain, M.~Pratama, E.~Lughofer, and S.G. Anavatti.
\newblock Evolving type-2 web news mining.
\newblock {\em Applied Soft Computing}, 54:200--220, 2017.

\bibitem{ZhangZhou14}
M.L. Zhang and Z.H. Zhou.
\newblock A review on multi-label learning algorithms.
\newblock {\em IEEE Transactions on Knowledge and Data Engineering},
  26(8):1819--1837, 2014.

\bibitem{Zhang04}
X.~Zhang.
\newblock {\em Matrix analysis and applications}.
\newblock Tsinghua University Press, 2004.

\bibitem{ZhangZhaoNiu19}
Y.~Zhang, P.~Zhao, S.~Niu, Q.~Wu, J.~Cao, J.~Huang, and M.~Tan.
\newblock Online adaptive asymmetric active learning with limited budgets.
\newblock {\em IEEE Transactions on Knowledge and Data Engineering},
  33(6):2680--2692, 2019.

\bibitem{ZhangYeung13}
Y.U. Zhang and D.Y. Yeung.
\newblock Multilabel relationship learning.
\newblock {\em ACM Transactions on Knowledge Discovery from Data}, 7(2):1--30,
  2013.

\bibitem{ZliobaiteBifetPfahringerHolmes14}
I.~Zliobaite, A.~Bifet, B.~Pfahringer, and B.~Holmes.
\newblock Active learning with drifting streaming data.
\newblock {\em IEEE Transactions on Neural Networks and Learning Systems},
  25(1):27--39, 2014.

\end{thebibliography}

\end{document}